\documentclass[lettersize,journal]{IEEEtran}
\usepackage{amsmath,amsfonts}
\usepackage{algorithmic}
\usepackage{algorithm}
\usepackage{array}
\usepackage[caption=false,font=normalsize,labelfont=sf,textfont=sf]{subfig}
\usepackage{textcomp}
\usepackage{stfloats}
\usepackage{url}
\usepackage{verbatim}
\usepackage{graphicx}
\usepackage{cite}
\usepackage{multirow}
\usepackage{array}
\usepackage{makecell}
\usepackage[hidelinks]{hyperref}
\hypersetup{
    colorlinks=true,
    linkcolor=black,
    filecolor=magenta,      
    urlcolor=black,
    citecolor=black,
    }
\begin{document}

\title{TopicFM+: Boosting Accuracy and Efficiency of Topic-Assisted Feature Matching}

\author{Khang Truong Giang, Soohwan Song*, Sungho Jo*
\thanks{* Corresponding author.}
\thanks{Khang Truong Giang and Sungho Jo are with School of Computing, KAIST, Daejeon 34141, Republic of Korea. Soohwan Song is with Intelligent Robotics Research Division, ETRI, Daejeon 34129, Republic of Korea. (Email addresses: \{khangtg,shjo\}@kaist.ac.kr, soohwansong@etri.re.kr).}}%

\markboth{}%
{Khang \MakeLowercase{\textit{et al.}}: TopicFM+: Boosting Accuracy and Efficiency of Topic-Assisted Feature Matching}


\maketitle

\begin{abstract}
This study tackles the challenge of image matching in difficult scenarios, such as scenes with significant variations or limited texture, with a strong emphasis on computational efficiency. Previous studies have attempted to address this challenge by encoding global scene contexts using Transformers. However, these approaches suffer from high computational costs and may not capture sufficient high-level contextual information, such as structural shapes or semantic instances. Consequently, the encoded features may lack discriminative power in challenging scenes. To overcome these limitations, we propose a novel image-matching method that leverages a topic-modeling strategy to capture high-level contexts in images. Our method represents each image as a multinomial distribution over topics, where each topic represents a latent semantic instance. By incorporating these topics, we can effectively capture comprehensive context information and obtain discriminative and high-quality features. Additionally, our method effectively matches features within corresponding semantic regions by estimating the covisible topics. To enhance the efficiency of feature matching, we have designed a network with a pooling-and-merging attention module. This module reduces computation by employing attention only on fixed-sized topics and small-sized features. Through extensive experiments, we have demonstrated the superiority of our method in challenging scenarios. Specifically, our method significantly reduces computational costs while maintaining higher image-matching accuracy compared to state-of-the-art methods.
\end{abstract}


\section{Introduction}
\IEEEPARstart{I}{mage} matching involves identifying corresponding pixels between two images at a detailed level. The outcomes of the matching process are crucial for various computer vision tasks, including structure from motion (SfM) \cite{schonberger2016structure,zhu2018very}, simultaneous localization and mapping (SLAM) \cite{mur2015orb,campos2021orb}, visual tracking \cite{tomasi1991tracking,shi1994good}, and visual localization \cite{sarlin2019coarse,lynen2020large}, etc. Traditional image-matching approaches \cite{lowe2004distinctive,bay2008speeded,calonder2011brief,sattler2012improving,jin2021image} generally consist of several steps: keypoint detection, keypoint description, keypoint matching, and outlier rejection. Typically, sparse keypoints are extracted using handcrafted local features, and matching is done through nearest neighbor search. More recently, convolutional neural networks (CNNs) have been utilized to extract local features \cite{yi2016lift,ono2018lf,detone2018superpoint}, surpassing the performance of conventional handcrafted features. However, these CNN-based methods often face challenges in scenarios with illumination variations, repetitive structures, or low-texture conditions, leading to a significant decline in their performance.

To address this problem, alternative methods that do not rely on detectors, \textit{detector-free methods}, have been suggested \cite{li2020dual,rocco2020efficient,rocco2018neighbourhood}. These methods enable direct feature matching on dense feature maps without the need for a keypoint detection stage. In particular, some studies \cite{sun2021loftr,wang2022matchformer,jiang2021cotr,chen2022aspanformer} have utilized Transformers \cite{dosovitskiy2020image} to learn robust feature representations. Transformers effectively capture comprehensive contextual information in images by iteratively employing self-attention and cross-attention layers. Consequently, they have achieved state-of-the-art results by generating numerous precise matches, even in cases involving repetitive patterns and textureless images.

\begin{figure}[t]
\centering
\includegraphics[width=3.5in]{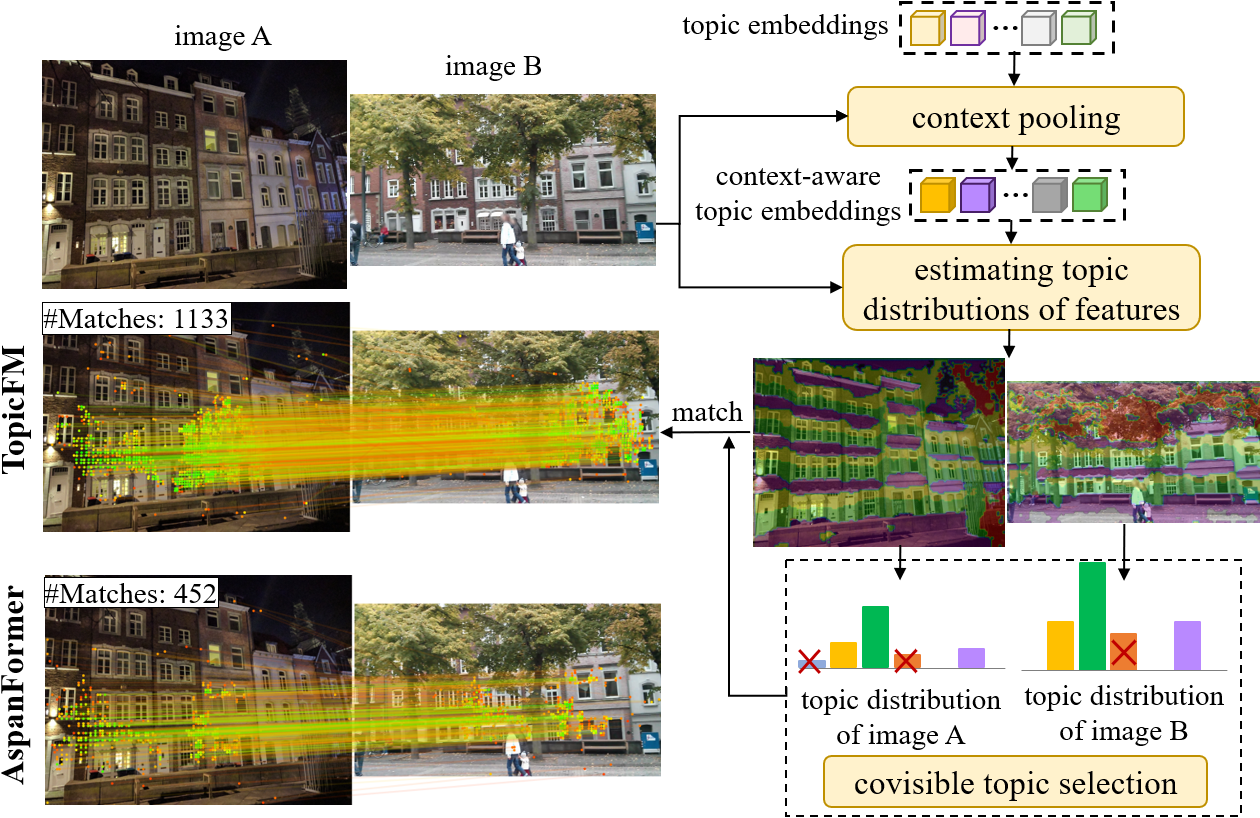} %
\caption{The main concept of our human-friendly topic-assisted feature matching, TopicFM. The method pools high-level context information from each image into a set of topics and then estimates a distribution over topics for each image feature. Consequently, the image is represented probabilistically with the topics marked in distinct colors. With this representation, TopicFM can efficiently identify overlapping regions by selecting covisible topics. By leveraging the selected topics and the unique information captured by each topic, TopicFM enhances the robustness of features, thus improving the matching accuracy.}
\label{fig_intro_1}
\end{figure}

However, despite the advantages of detector-free methods, they still encounter several challenges that hinder the feature-matching performance: (i) These methods struggle to sufficiently incorporate the global context of a scene, which is crucial for accurate feature matching. Some approaches \cite{sun2021loftr,wang2022matchformer,chen2022aspanformer} have attempted to implicitly capture global contextual information using Transformers. However, higher-level contexts, such as semantic instances or structural shapes, should be explicitly exploited to learn robust representations. (ii) These methods exhaustively search for features across the entire image area. This can lead to interference from keypoints in non-overlapping regions, resulting in incorrect matches and increased computation time. As a result, their matching performance significantly suffers when there are limited areas of overlap between images. (iii) Existing detector-free methods \cite{chen2022aspanformer,sun2021loftr,dai2023oamatcher} also demand substantial computational resources and extensive memory for dense matching. Consequently, they are limited to low-resolution images or pose challenges in real-time applications. Therefore, minimizing the runtime and GPU memory usage becomes crucial to deploy detector-free methods on devices with memory and computational limitations. (iv) Several studies \cite{zhou2021patch2pix,sun2021loftr,wang2022matchformer} have adopted a coarse-to-fine strategy to handle high-resolution images. This approach involves initially finding matches at a coarse level and refining them at a finer level. However, keypoints identified at the coarse level may lack sufficient information, and they can end up being located in occluded or non-overlapping regions at the fine level, making it difficult to determine accurate matching locations.

\begin{figure}[!t]
\centering
\includegraphics[width=3.48in]{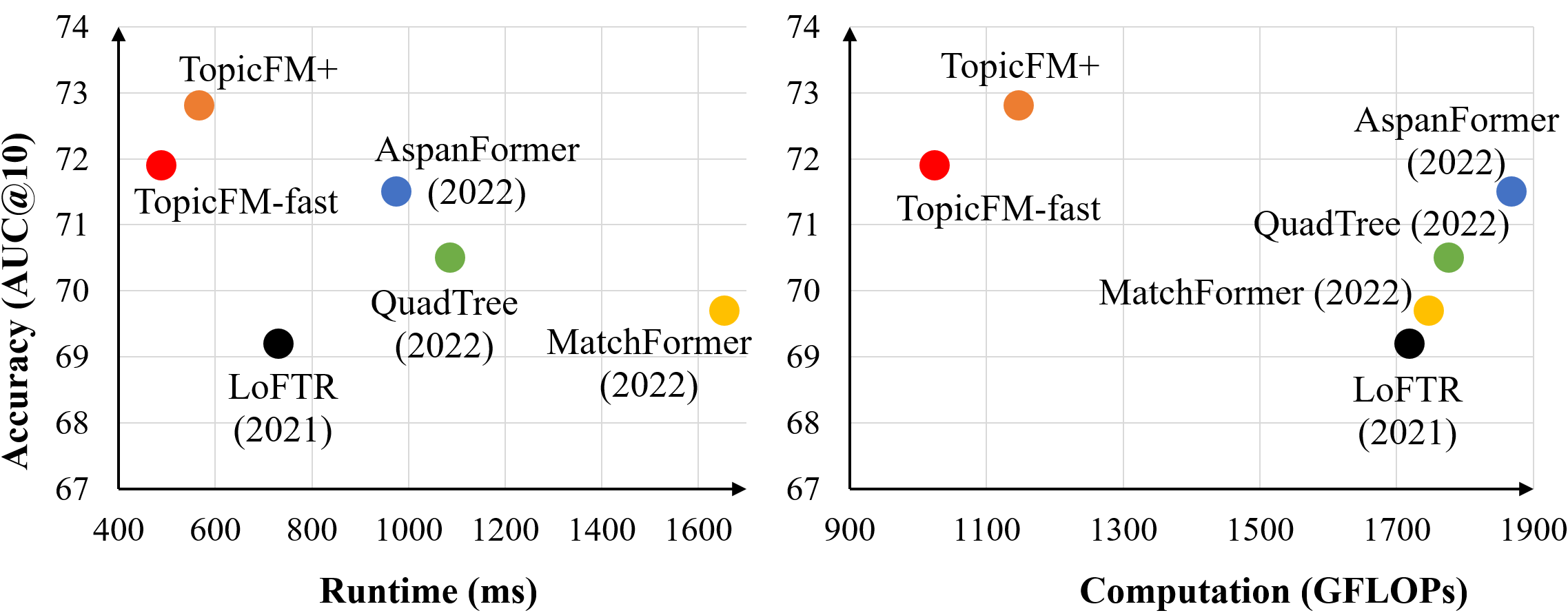}
\caption{Comparison between the proposed models (TopicFM-fast, TopicFM+) and SOTA Transformer-based methods on the MegaDepth dataset \cite{li2018megadepth}. We report accuracy (AUC@$10^o$), runtime (ms), and computational cost (GFLOPs). The runtime and computational cost are measured at the image resolution of $1216 \times 1216$}
\label{fig_intro_2}
\end{figure}

To address these issues, we propose a robust and efficient method for detector-free feature matching called \textbf{TopicFM}. This method incorporates high-level contextual information from images using a \textit{topic modeling strategy} \cite{blei2003latent,yan2013biterm} inspired by data mining techniques. In TopicFM, an image is represented as a multinomial distribution over topics, with each topic representing a latent semantic instance like an object or structural shape. By leveraging this representation, TopicFM performs probabilistic feature matching based on the distribution of the latent topics. Fig. \ref{fig_intro_1} illustrates the representative topics inferred from the image-matching results. The image regions with the same semantic instance or shape are assigned to the same topic. Through the comprehensive context information captured by the topics, TopicFM can learn discriminative and high-quality features. Furthermore, TopicFM effectively matches features within overlapping regions of image pairs by estimating the topics that are covisible between them. This approach resembles the cognitive process of humans, who quickly recognize covisible regions based on semantic information and then search for matching points within those regions. By applying this top-down approach, our method successfully identifies robust and accurate matching points, even in challenging cases involving significant variations in scale and viewpoint.

Moreover, to further improve computational performance, we have developed an efficient end-to-end network architecture based on a coarse-to-fine framework. In the coarse-level matching stage, we introduce a pooling-and-merging attention network in TopicFM to streamline computation. This network comprises two crucial steps: context pooling and context merging. The pooling step captures high-level contexts using latent topics, while the merging step utilizes these topics to enhance feature robustness. Both steps have low computational costs as they involve attention layers with small-sized inputs. Unlike other transformer-based methods that require computationally expensive multiplications between large feature matrices, our attention layers only operate on smaller feature matrices or fixed-sized topic matrices. Fig. \ref{fig_intro_2} illustrates the impressive achievements of our method, which achieves an approximately 50\% reduction in both runtime and computational cost while surpassing the state-of-the-art Transformer-based methods in image-matching accuracy.

In the fine-level matching stage, we improve upon coarse matches by determining an adaptive feature location rather than relying on a fixed center location \cite{sun2021loftr,chen2022aspanformer}. To achieve this, we propose a dynamic feature refinement module that employs a self-feature-detector trained using a symmetric epipolar distance loss \cite{hartley2003multiple}. This loss function does not require explicit ground-truth feature correspondences, allowing the learned network to automatically prioritize easily matchable points, such as peak points or overlapping points, as the keypoints of interest. By utilizing these self-detected keypoints instead of a center-fixing keypoint approach, we observe an overall enhancement in matching accuracy. To ensure efficiency in the fine-level matching process, we design the self-detector network using MLP-Mixer layers, which are more computationally efficient than transformer layers \cite{tolstikhin2021mlp}. As a result, our coarse-to-fine method produces high-quality matching results while requiring less computation compared to state-of-the-art transformer-based methods \cite{sun2021loftr,chen2022aspanformer,tang2022quadtree}.
The contributions of our study can be summarized as follows:
\begin{itemize}
    \item We introduce a novel feature-matching approach that integrates local context and high-level semantic information into latent features using a topic modeling strategy. By inferring covisible topics, this method achieves accurate dense matches in challenging scenes.
    \item We formulate the topic inference process as a learnable transformer module, enabling interpretability of the matching results, even from a human perspective.
    \item To enhance fine-level matching performance, we propose dynamic feature refinement, which identifies highly matchable features within image patches. This refinement process is trained without explicit reliance on ground-truth matches, allowing the automatic selection of reliable keypoints.
    \item We have designed an efficient end-to-end network that reduces dependence on the Transformer layer. The network leverages attention mechanisms operating on fixed-sized latent topics and small-sized image features in the coarse stage, and utilizes an efficient MLP-Mixer network in the fine stage. This model considerably accelerates the processing of image frames.
    \item We extensively evaluate the robustness and efficiency of the proposed method through empirical experiments. Additionally, we demonstrate the interpretability of our topic models. The source code for our method is available at \url{https://github.com/TruongKhang/TopicFM}
\end{itemize}

A preliminary version of this paper has been previously presented in \cite{khang2023topicfm}. In this study, we introduce an upgraded and more efficient approach for the coarse-level matching step by utilizing the pooling-and-merging attention module. While the pooling step performs topic inference as described in \cite{khang2023topicfm}, the merging step enhances the features by directly merging them with the inferred topic embeddings. This merging step eliminates the need for in-topic feature augmentation, as presented in \cite{khang2023topicfm}. The previous augmentation step involved employing attention layers for all features within a topic, which incurs a higher computation cost compared to the merging step proposed in this paper. Additionally, we introduce a novel dynamic feature refinement network in the fine-level matching step, which is trained in a self-supervised manner relying solely on the ground-truth fundamental matrices. This network enhances the robustness of the matching process at a finer level. Furthermore, we provide comprehensive validation of our methods through additional experiments, evaluations, and detailed analyses. These efforts aim to demonstrate both the effectiveness and efficiency of the proposed approach.

\section{Related Works}
\label{related_works}
\subsection{Image Matching}
\label{related_works_image_matching}
Image matching has long been a prominent challenge in the field of computer vision. Typically, two main approaches are employed: detector-based and detector-free feature matching. The detector-based approach involves using traditional algorithms \cite{lowe2004distinctive,bay2008speeded,rosten2006machine} or deep CNNs \cite{noh2017large,ono2018lf,dusmanu2019d2,detone2018superpoint} to detect a set of keypoints of interest in each input image. These keypoints are then used to extract high-dimensional feature vectors by encoding their contextual information. The context can be represented by image patches \cite{lowe2004distinctive,bay2008speeded} or implicitly captured through various types of CNNs \cite{revaud2019r2d2,bhowmik2020reinforced,tyszkiewicz2020disk}. Once the features of interest are obtained for each image, the matching process is usually performed for a pair of images by computing pairwise similarity between the two sets of features. Subsequently, mutual nearest neighbor search \cite{muja2014scalable} or ratio test \cite{lowe2004distinctive} is applied to establish correspondences.

However, these methods primarily consider local context information, neglecting the global or geometric connections between features. Consequently, the extracted features are less effective when dealing with challenging cases. To address this limitation, some studies \cite{luo2019contextdesc,luo2020aslfeat,sarlin2020superglue,chen2021learning,shi2022clustergnn} have incorporated global context information to improve the distinctiveness of detected features. ContextDesc \cite{luo2019contextdesc} and ALSFeat \cite{luo2020aslfeat} introduced a geometric context encoder using a large patch sampler and a deformable CNN, respectively. SuperGlue \cite{sarlin2020superglue} and SGMNet \cite{chen2021learning} utilized self- and cross-attentions of a graph neural network to integrate global visual and geometric information into local feature representations. While these detector-based methods \cite{sarlin2020superglue,luo2019contextdesc,luo2020aslfeat,detone2018superpoint} are efficient due to sparse matching, their performance heavily relies on the separated keypoint detection step, which still struggles in challenging conditions such as small overlapped regions or large untextured scenes.

To address this issue, researchers have introduced detector-free methods \cite{zhou2021patch2pix,sun2021loftr,wang2022matchformer,chen2022aspanformer}. These methods have designed end-to-end network architectures that perform image matching in a single forward pass, eliminating the need for separate steps. Instead of extracting sparse feature points, these networks directly process dense feature maps. To efficiently handle the dense features of high-resolution images, many studies have employed a coarse-to-fine strategy. Patch2Pix \cite{zhou2021patch2pix} detects patch-level coarse matches in low-resolution images and progressively refines them at higher resolutions. Similarly, several coarse-to-fine methods \cite{sun2021loftr,wang2022matchformer,jiang2021cotr,chen2022aspanformer} utilize transformers to learn robust and distinctive features. LoFTR \cite{sun2021loftr} employs a linear transformer \cite{shen2021efficient} to enhance the representation ability of visual descriptors by incorporating global context information. Matchformer \cite{wang2022matchformer} is an extract-and-match scheme that intuitively combines self- and cross-attention in each stage of the hierarchical encoder. This match-aware encoder reduces the burden on the decoder, resulting in a highly efficient model. QuadTree \cite{tang2022quadtree} constructed token pyramids similar to quadtree data structures and calculates attention in a step-by-step manner, starting from coarse regions and gradually refining to finer regions. This approach effectively reduces the computational complexity by progressively eliminating irrelevant subdivided rectangular regions. ASpanFormer \cite{chen2022aspanformer} utilized a hierarchical attention framework for feature matching, incorporating attention operations at multiple scales to facilitate both global context understanding and precise matching. The span of attention is dynamically determined, allowing the model to capture both long-range dependencies and fine-grained details in local regions.

However, these existing methods \cite{sun2021loftr,wang2022matchformer,tang2022quadtree,chen2022aspanformer} face inefficiencies when propagating global context information across the entire image region. We argue that the unobserved regions between image pairs are unnecessary and may introduce noise when utilizing transformers for feature learning. Additionally, as previously mentioned, transformers demand substantial computational resources and memory, making them unsuitable for real-time processing or lightweight models. To address these issues, we propose a topic modeling approach that leverages relevant contextual cues for feature learning. Our approach involves classifying features into latent topics and utilizing the distinctive context information encoded within each topic to improve the quality of features. This strategy effectively reduces computation by applying attention layers exclusively on inputs with limited sizes. We demonstrate that our design achieves state-of-the-art performance while being remarkably more efficient than existing methods.

\begin{figure*}[!t]
\centering
\includegraphics[width=7.0in]{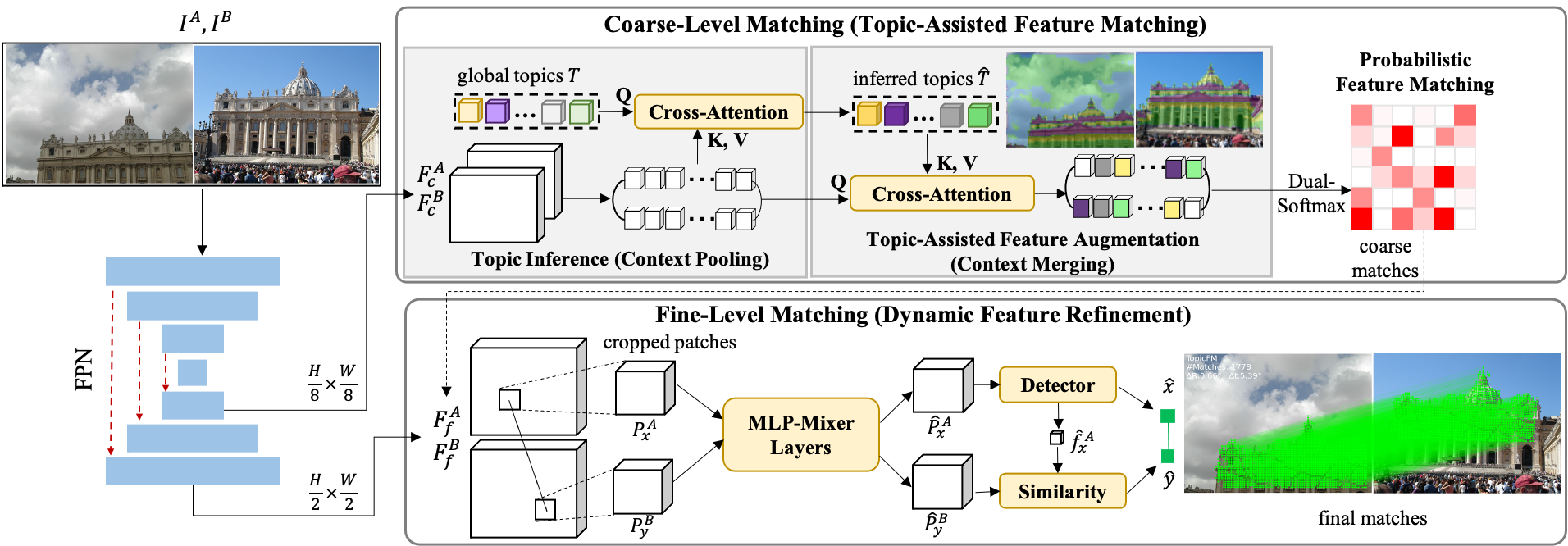}
\caption{The architecture of proposed image-matching method. We design a Feature Pyramid Network (FPN) that extracts feature maps at the low  $\left(\frac{H}{8} \times \frac{W}{8}\right)$ and high $\left(\frac{H}{2} \times \frac{W}{2}\right)$ resolutions. Building upon this, we introduce a \textbf{topic-assisted feature matching} module (Section \ref{method_topicfm}) and a \textbf{dynamic refinement network} (Section \ref{method_dynamic_refinement}) to perform coarse-level and fine-level matching, respectively.}
\label{fig_method_architecture}
\end{figure*}

\subsection{Interpretable Image Matching}
\label{related_works_interpretability}
The interpretability has emerged as a prominent area of research in the field of computer vision models. Its primary objective is to provide explanations for the decisions and predictions made in image recognition \cite{zhang2018interpretable,zhang2019interpreting,williford2020explainable} and deep metric learning \cite{zhao2021towards}. In the context of image matching, detector-based methods have been employed to estimate interpretable feature keypoints such as corners, blobs, or ridges. However, these detected features fail to capture the spatial or semantic structures present in the images. On the other hand, existing end-to-end methods rely on CNNs \cite{zhou2021patch2pix} or transformers to extract dense feature maps using either local or global context \cite{sun2021loftr,wang2022matchformer}. Nonetheless, these approaches do not explicitly convey the specific details of the observed context, limiting the interpretability of their outcomes.

In contrast, the human cognitive system rapidly identifies covisible regions by leveraging high-level contextual information, such as objects or structures, and subsequently determines the matching points within these regions. Drawing inspiration from this cognitive process, several studies \cite{jiang2021cotr,chen2022guide,dai2023oamatcher} have proposed image-matching methods that explicitly extract covisible regions. These methods predict covisible regions of two images, such as overlapping bounding boxes \cite{chen2022guide} or overlapping masks \cite{jiang2021cotr,dai2023oamatcher}, and conduct feature matching exclusively within these regions. However, these approaches entail a significant computational overhead in finding covisible regions and lack sufficient interpretability for users. Moreover, inaccurate identification of covisible regions can substantially impact the quality of matching as these methods directly perform feature matching within the covisible region.

On the other hand, our method efficiently categorizes local structures within images into distinct topics and implicitly harnesses the information within these topics to enhance the features. Instead of employing binary labels, we model an image as a multinomial distribution of topics and subsequently perform probabilistic feature matching. Additionally, our method ensures interpretable matching by selecting crucial topics within the covisible regions of the two images. To the best of our knowledge, our method is the first to introduce a high level of interpretability to the task of image matching.

\subsection{Transformers with Attention}
\label{related_works_transformers}
The Transformer architecture gained popularity in various vision tasks during the early 2020s and demonstrated state-of-the-art performance in image classification tasks \cite{dosovitskiy2020image,liu2021swin}. However, when it comes to feature matching for camera pose estimation, there is a significant difference compared to classification. In feature matching, the spatial location of features plays a crucial role in capturing precise context information and learning distinctive features across images. Although the Transformer layer uses positional encoding to incorporate the spatial information of features, it cannot effectively leverage this information to observe adequate context for feature learning. In fact, most existing Transformer-based feature-matching methods \cite{sun2021loftr,chen2022aspanformer,dai2023oamatcher,yu2023adaptive} still rely on a deep CNN backbone to extract the initial features. This raises the question of whether extensive self/cross attention between the features is necessary for the feature-matching task. Our proposed method addresses this question by applying attention only to the small-length features and the fixed-sized latent topics, demonstrating that such an approach is sufficient for achieving effective feature matching.

\section{Proposed Method}
\label{method}
The goal of image matching is to establish correspondences between features extracted from two images, represented as feature maps $\{F^A,F^B\}$ derived from images $\{I^A,I^B\}$. Specifically, it aims to determine a set of feature correspondences $\{(f_i^A,f_j^B)\}$, where $f_i^A \in F^A$ and $f_j^B \in F^B$. To achieve accurate pixel-level matching while ensuring computational efficiency, we have adopted a coarse-to-fine approach. This approach involves initially estimating coarse matches at a lower image resolution and then refining them at a higher resolution. The overall architecture, depicted in Fig. \ref{fig_method_architecture}, comprises three main components: feature extraction, coarse-level matching (Section \ref{method_topicfm}), and fine-level matching (Section \ref{method_dynamic_refinement}).

In the feature extraction stage, we employ a feature pyramid network (FPN) \cite{lin2017feature} to generate multi-scale feature maps. The multi-scale feature extraction network is constructed based on an efficient standard CNN block, which comprises a Conv2D layer, BatchNorm layer, and GELU activation layer. This choice of design improves computational efficiency significantly compared to using a ResNet block \cite{he2016deep}. For the coarse and fine stages, we choose feature maps with resolutions of $\frac{H}{8} \times \frac{W}{8}$ and $\frac{H}{2} \times \frac{W}{2}$, respectively, denoted as $\{F_c^A,F_c^B\}$ and $\{F_f^A, F_f^B\}$, where H and W represent the height and width of the original image.

In the coarse-level matching stage, we introduce topic-assisted feature matching to estimate the matching distribution $P(M)$ between the coarse feature maps $F_c^A$ and $F_c^B$. This feature matching module is designed as a pooling-and-merging attention network, which shares similarities with the squeeze-and-excitation concept \cite{8578843}. This network consists of two essential steps: context pooling and context merging. In the context pooling step, we use cross-attention to infer high-level contexts, topics, for each image. Then, in the context merging step, we leverage these inferred topics to enhance the robustness of features. The fixed size of the latent topics allows us to achieve both accurate estimation of the matching distribution P(M) and computational efficiency. With the computed distribution, we determine a set of coarse matches $\{(x,y)\}$ $(x \in F_c^A,y \in F_c^B)$ by extracting feature pairs with high matching confidence and satisfying a mutual nearest neighbor condition.

In the fine-level matching stage, we upscale the coarse matching coordinates to the high-resolution feature maps $\{F_f^A,F_f^B\}$.  Then, we crop image patches $\{(P_x^A,P_y^B)\}$ centered around these upscaled coordinates. To detect the final matches $\{(\Hat{x}, \Hat{y})\}$, where $\Hat{x} \in P_x^A$ and $\Hat{y} \in P_y^B$, we propose a dynamic refinement network that operates on these cropped patches. Unlike existing methods \cite{sun2021loftr,wang2022matchformer,chen2022aspanformer} that use a fixed center location for each patch, our approach identifies more reliable keypoint locations by estimating a score map for each patch. Importantly, this network is trained using a self-supervised learning framework, eliminating the need for ground-truth matching pairs. This enables the detection of reliable keypoints and enhances the adaptability and robustness of the feature refinement process.

\subsection{Topic-Assisted Feature Matching}
\label{method_topicfm}
Given a pair of feature maps $\{F_c^A,F_c^B\}$ extracted from an image pair $\{I^A,I^B\}$ at the coarse level, our objective is to estimate the matching distribution $M=\{m_{ij}\}$ \cite{bhowmik2020reinforced}:
\begin{equation}
\label{eq:prob_feat_match}
    P(M \mid F_c^A,F_c^B ) = \prod_{m_{ij} \in M} P \left( m_{ij} |F_c^A,F_c^B \right)
\end{equation}
where the binary random variable $m_{ij}$ indicates that the $i^{th}$ feature $F_{c,i}^A$ is matched to the $j^{th}$ feature $F_{c,j}^B$. Previous methods \cite{sarlin2020superglue,sun2021loftr} have directly employed attention-based networks on all feature tokens to estimate the distribution of $M$, followed by the use of Dual-Softmax \cite{sun2021loftr} or optimal transport layers \cite{sarlin2020superglue,cuturi2013sinkhorn}. However, this approach leads to high computational overhead, especially when dealing with a large number of features. Additionally, applying attention to all features can introduce noise from non-overlapping regions, resulting in degraded performance. To tackle these challenges, we incorporate latent topics to encode sufficient and semantic context information into the features. These latent topics capture high-level contexts, such as objects or structural shapes, in the images. By understanding the underlying topics within each image, we can improve the matching probability while mitigating the computational complexity associated with existing methods.

To implement this idea, we take inspiration from topic modeling techniques used in data mining \cite{blei2003latent}, where each document in a corpus consists of $K$ latent topics, and each topic is characterized by a collection of tokens that share the same semantic information. In the context of feature matching, we can consider each image as a document and each feature point as a word. Consequently, the image can be represented as a multinomial distribution of $K$ topics, where each topic encompasses a set of feature "tokens" that capture high-level contextual information. Our goal is to estimate a distribution over topics (referred to as topic distribution) for each feature point. We denote the topic indicator of feature $f_i$ as $z_i$, where $z_i$ belongs to the set $\{1,2,…,K\}$, and the topic distribution for $f_i$ as $\theta_i \in \mathbb{R}^K$, where $\theta_{i,k}$ represents the probability $p(z_i=k)$.

To incorporate latent topics into Eq. \ref{eq:prob_feat_match}, we introduce the random variable $z_{ij}$, which represents the assigned topic for a pair of feature points $\{F_{c,i}^A, F_{c,j}^B\}$. Here, $z_{ij}$ can take values from the set $\mathcal{Z}=\{1,2,…,K,NaN\}$. If $z_{ij}=k$, it indicates that $F_{c,i}^A$ and $F_{c,j}^B$ are co-assigned to topic $k$. Conversely, if $z_{ij}$  is $NaN$, the feature pair does not belong to the same topic and is considered highly unmatchable. Eq. \ref{eq:prob_feat_match} can be rewritten as follows:
\begin{multline}
\label{eq:log_likelihood}
    \log P\left(M \mid F_c^A, F_c^B \right) = \sum_{m_{ij} \in M} \log P  \left(m_{ij} \mid F_c^A, F_c^B \right) \\
    = \sum_{m_{ij} \in M} \log \sum_{z_{ij} \in \mathcal{Z}} P \left(m_{ij}, z_{ij} \mid F_c^A, F_c^B \right)
\end{multline}
To compute Eq. \ref{eq:log_likelihood}, we employed an evidence lower bound (ELBO) function:
\begin{multline}
\label{eq:lower_bound}
    \mathcal{L}_{ELBO} 
    = \sum_{m_{ij} \in M} \sum_{z_{ij} \in \mathcal{Z}} P\left(z_{ij} \mid F_c \right) \log P\left(m_{ij} \mid z_{ij}, F_c \right) \\
    = \sum_{m_{ij}} E_{p(z_{ij})}  \log P\left(m_{ij} \mid z_{ij}, F_c\right)
\end{multline}
where $P(m_{ij} |z_{ij},F_c^A,F_c^B)$ represents the conditional matching probability of the match $m_{ij}$ given the co-assigned topic $z_{ij}$, which contains semantic contextual information. Eq. \ref{eq:lower_bound} can be computed using Monte-Carlo (MC) sampling as follows:
\begin{gather}
\label{eq:conditional_match_dis}
    \mathcal{L}_{ELBO} = \sum_{m_{ij}  \in M} \frac{1}{S} \sum_{s=1}^S \log P\left(m_{ij} \mid z_{ij}^{(s)}, F_c^A, F_c^B \right) \\
    z_{ij}^{(s)} \sim P\left(z_{ij} \mid F_c^A, F_c^B \right)
\label{eq:topic_sampling}
\end{gather}
Based on Eqs. \ref{eq:conditional_match_dis} and \ref{eq:topic_sampling}, our objective is to infer the topic distribution $P(z_{ij} |F_c^A,F_c^B)$  and then estimate the topic-assisted matching distribution $P(m_{ij} |z_{ij}^{(s)},F_c^A,F_c^B)$. The topic inference step involves grouping features that share similar global contexts into topics and representing these contexts using topic embedding vectors. This process is referred to as context pooling. Subsequently, we utilize the inferred topics to enhance the features, where features with similar topic distributions are represented more closely. We perform topic-assisted feature augmentation to incorporate topic information into the features. This step can be understood as context merging. After obtaining the updated features, we calculate the matching probabilities to extract coarse correspondences. In the following subsections, we will provide a comprehensive and detailed explanation of this process..

\subsubsection{Topic Inference -- Context Pooling}
\label{method_context_pooling}
The distribution of $z_{ij}$ is inferred by decomposing it into two separate distributions, $z_i$ and $z_j$. This decomposition is represented as follows:
\begin{equation}
\label{eq:topic_dis_k}
\textstyle P(z_{ij} = k \mid F_c) = \\ P(z_i = k \mid F_c^A )  P(z_j = k \mid F_c^B) = \theta_{i,k}^A \theta_{j,k}^B
\end{equation}
where $\theta_i^A$ and $\theta_j^B$ are the topic distributions of feature $F_{c,i}^A$ and $F_{c,j}^B$, respectively. Eq. \ref{eq:topic_dis_k} demonstrates that the distribution of the topic co-assignment can be estimated by individually inferring the topic distribution for each feature point.

To perform individual topic inference, we utilized cross-attention. We introduced a learnable global representation of topic $k$, denoted as $T_k \in \mathbb{R}^D$. Through cross-attention layers, we estimated the local representation $\Hat{T}$ conditioned on the coarse feature map $F_c$ as follows:
\begin{equation}
\label{eq:context_pooling}
    \Hat{T}_k = \texttt{CA}(T_k, F_c)
\end{equation}
where $\texttt{CA}(T_k,F_c)$  represents an attention layer with queries $T_k$, keys $F_c$, and values $F_c$. Eq. \ref{eq:context_pooling} serves two purposes: i) categorizing different spatial structures or objects within the feature map $F_c$ into distinct topics, and ii) representing these structures with updated topic embeddings $\Hat{T}$. Geometrically, $\Hat{T}_k$ can be considered as the centroid of the shapes and semantic structures contained in topic $k$. 

Next, we applied the Softmax layer to estimate the topic distribution $\theta_i$ for each feature $f_i \in F_c$, as follows:
\begin{equation}
\label{eq:topic_infer_norm}
    \theta_{i, k} = \frac{\exp\left(\langle \hat{T}_k, F_i \rangle\right)}{\sum_{h=1}^K \exp\left(\langle \hat{T}_h, F_i \rangle\right)}
\end{equation}
where $\langle.,.\rangle$ represents the dot-product operator. After performing individual topic inference in Eq. \ref{eq:topic_infer_norm}, we computed the probability of co-assigning the feature pair $\{F_{c,i}^A, F_{c,j}^B\}$ to topic $k$ following Eq. \ref{eq:topic_dis_k}. However, obtaining the explicit topic label $k$ for each feature pair during training presents a challenge. Therefore, we defined the probability of co-assigning the feature pair to an implicit topic as follows:
\begin{equation}
\label{eq:topic_dis_all_k}
    P(z_{ij} \in \{1…K\} \mid F_c^A,F_c^B) = \sum_{k=1}^K \theta_{i,k}^A \theta_{j,k}^B = \langle\theta_i^A, \theta_j^B\rangle
\end{equation}
Additionally, the probability of the feature pair not belonging to the same topic was calculated as follows:
\begin{equation}
\label{eq:topic_dis_nan}
    P(z_{ij} = NaN \mid .) = 1 - P(z_{ij} \in \{1...K\} \mid .) = 1 - \langle\theta_i^A, \theta_j^B\rangle
\end{equation}
During training, given the ground-truth feature pairs, Eqs. \ref{eq:topic_dis_all_k} and \ref{eq:topic_dis_nan} are considered topic-matching formulas used to train the latent topics.

\subsubsection{Topic-Assisted Feature Augmentation -- Context Merging}
\label{method_context merging}
After the topic inference step, topic labels were sampled for each feature $f_i$ based on the estimated topic distribution $\theta_i$. The features were then clustered into their respective topics. We denote the sets of features belonging to topic $\Tilde{k} = z_{ij}^{(s)}$ as $F_c^{A,\Tilde{k}} \subset F_c^A$ and $F_c^{B,\Tilde{k}} \subset F_c^B$. These features capture the high-level contexts associated with topic $\Tilde{k}$. Then, self/cross-attention layers were applied to encode those contexts into the features:
\begin{equation}
\label{eq:in_topic_feature_augmentation}
\begin{split}
    F_{c}^{A, \Tilde{k}} &= \texttt{SA} \left( F_{c}^{A, \Tilde{k}}, F_c^{A, \Tilde{k}} \right),
    F_{c}^{B, \Tilde{k}} = \texttt{SA} \left( F_{c}^{B, \Tilde{k}}, F_c^{B, \Tilde{k}} \right) \\
    F_{c}^{A, \Tilde{k}} &= \texttt{CA} \left( F_{c}^{A, \Tilde{k}}, F_c^{B, \Tilde{k}} \right),
    F_{c}^{B, \Tilde{k}} = \texttt{CA} \left( F_{c}^{B, \Tilde{k}}, F_c^{A, \Tilde{k}} \right) \\
\end{split}
\end{equation}
We selectively focused on co-visible topics between the image pair for feature augmentation to improve computational efficiency. Co-visible topics were determined by comparing the topic distributions of the two images. The topic distribution of an image is estimated by integrating the distributions of all features:
\begin{equation}
\label{eq:covisible_topic_selection}
    \theta_k^A \propto \sum_{i=1}^{|F_c^A|} \theta_{i,k}^A, \quad \theta_k^B \propto \sum_{j=1}^{|F_c^B|} \theta_{j,k}^B
\end{equation}
where $\propto$ denotes the normalization operator. The co-visible probability was then computed by multiplying the two image-level topic distributions, denoted as $\theta^{covis}=\theta^A \theta^B$. Finally, the topics with the highest probabilities were selected for feature augmentation.

\textbf{Efficiency Improvement Strategy}. Although the feature-augmentation algorithm above is computationally efficient compared to Transformer-based methods \cite{sun2021loftr,chen2022aspanformer,wang2022matchformer,dai2023oamatcher}, it encounters a bottleneck when most of the features are assigned to a single topic. This limitation arises from large textureless scenes or noise during topic inference. To address this issue and further enhance efficiency, we propose an upgraded version of feature augmentation. In this upgraded approach, we eliminated the topic sampling and the in-topic self-/cross-attention (Eq. \ref{eq:in_topic_feature_augmentation}). Instead, after topic inference, the features are merged directly with the context-aware topic embeddings $\Hat{T}$ using cross-attention layers, as follows:
\begin{equation}
\label{eq:efficient_context_merging}
    \Hat{F}_c = \texttt{CA}\left(F_c, \Hat{T}\right)
\end{equation}
where $\Hat{F}_c$ are the enhanced features after this context merging step. It is observed that the attention layer in Eq. \ref{eq:efficient_context_merging} estimated the amount of context information received for each feature point $F_{c,i}$ as follows:
\begin{equation}
\label{eq:expectation_amount_context}
    E_i[\Hat{T}] = \sum_{k=1}^K \theta_{i,k} \Hat{T}_k
\end{equation}
where $\theta_i$ is the topic distribution of feature $F_{c,i}$ estimated in Eq. \ref{eq:topic_infer_norm}. Eq. \ref{eq:expectation_amount_context} indicates that if two features have a similar topic distribution $\theta$ (i.e., belong to the same topic), they receive the same amount of context information and are highly matchable. This demonstrates the effective assistance of inferred topics in learning distinctive features across images. Notably, Eq. \ref{eq:efficient_context_merging} updates the features from fixed-sized topic matrix $\Hat{T}$, in which the number of topics $K$ is much smaller than the number of features assigned to topic $\Tilde{k}$ $(F_c^{A,\Tilde{k}}, F_c^{B,\Tilde{k}})$.Therefore, using Eq. \ref{eq:efficient_context_merging} for feature augmentation is more efficient than using Eq. \ref{eq:in_topic_feature_augmentation}. However, we find that Eq. \ref{eq:efficient_context_merging} slightly decreases the robustness of features compared to Eq. \ref{eq:in_topic_feature_augmentation}, leading to a minor reduction in performance, as described in the ablation study (Section \ref{ablation_study}).

\subsubsection{Coarse Match Estimation}
\label{method_coarse_match_est}
After the context merging step, let $\Hat{F}_c^A$ and $\Hat{F}_c^B$ be the augmented versions of $F_c^A$ and $F_c^B$, respectively. The dual-softmax operation \cite{sun2021loftr} was applied to compute the matching probabilities as follows: 
\begin{equation}
\label{eq:dual_softmax}
    P(m_{ij} \mid z_{ij}^{(s)} = k, F_c) = \texttt{DualSoft}\left(\langle \Hat{F}_{c,i}^A, \Hat{F}_{c,j}^B \rangle\right)
\end{equation}
Using Eq. \ref{eq:dual_softmax}, we obtain a matching confidence matrix $P_c \in [0, 1]^{N^A \times N^B}$, where $N^A$ and $N^B$ represent the lengths of $F_c^A$ and $F_c^B$, respectively. Finally, a confidence threshold $\tau$ and a mutual nearest neighbor search \cite{muja2014scalable} were applied to extract coarse matches.

\begin{figure}[!t]
\centering
\includegraphics[width=3in]{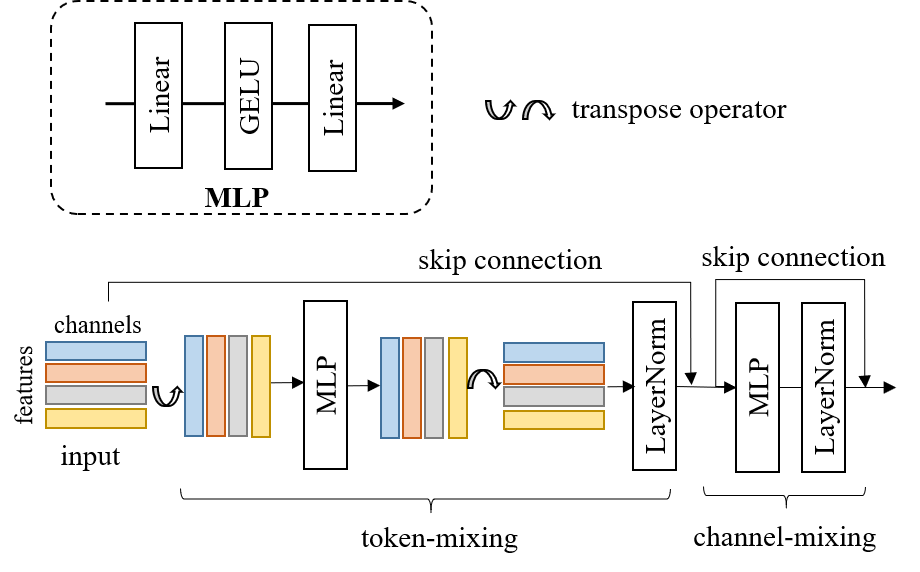}
\caption{Details of the MLP-Mixer block. This block extracts new features by employing MLP layers on both the spatial side (token-mixing) and the channel side (channel-mixing) independently.}
\label{mlp_mixer_block}
\end{figure}

\subsubsection{Training Latent Topics}
\label{method_training_topics}
We devised an effective procedure for learning latent topics. To mitigate topic overfitting, we employed a dropout layer on the initial topic embeddings $T$ before the context pooling step. This regularization technique prevents the dominance of a few topics during training and prediction. Then, we utilized the topic-matching formulas mentioned in Eqs. \ref{eq:topic_dis_all_k} and \ref{eq:topic_dis_nan}.

For a given ground truth matching pair $(i,j)$ at the coarse stage, we enforced the feature vectors $(f_i^A,f_j^B)$ to belong to the same topic. To achieve this, the negative log-likelihood of Eq. \ref{eq:topic_dis_all_k} was calculated as follows: 
\begin{equation}
    \mathcal{L}_{ij}^m = -\log \left( (\theta_i^A)^T \theta_j^B \right)
\end{equation}
We also introduced a non-matching loss to prevent assigning all features to a single topic. For each ground truth match $(i,j)$, we sampled $N$ unmatched pairs $\{(i,n)\}_{n=1}^N$ and defined the loss using Eq. \ref{eq:topic_dis_nan}: 
\begin{equation}
    \mathcal{L}_{in}^u = -\log \left( 1 - (\theta_i^A)^T \theta_n^B \right)
\end{equation}
The final loss, known as the topic matching loss, combines the two losses above as follows:
\begin{equation}
    \mathcal{L}_c^{topic} = \frac{1}{|M_c^{gt}|} \sum_{(i,j) \in M_c^{gt}} \mathcal{L}_{ij}^m + \frac{1}{|N * M_c^{gt}|} \sum_{(i, n)} \mathcal{L}_{in}^u 
\end{equation}
where $M_c^{gt}$ represents the number of ground truth matching pairs.

\begin{figure}[!t]
\centering
\includegraphics[width=3in]{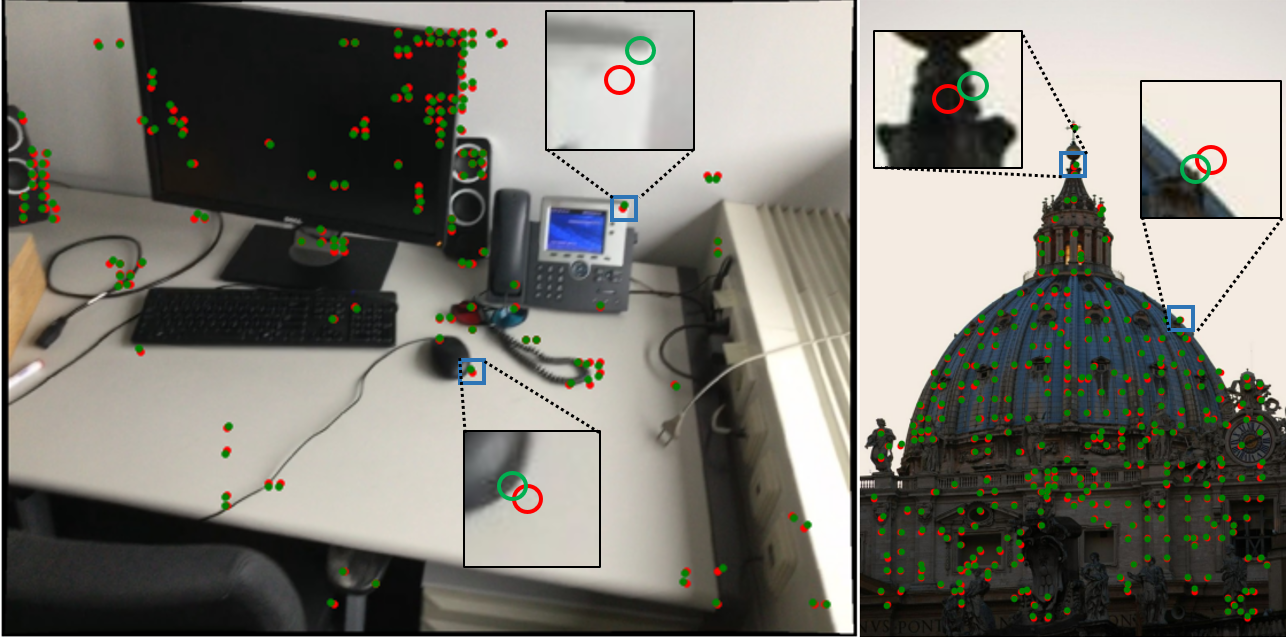}
\caption{Visualization of dynamic refinement results. The keypoints estimated in the coarse stage are highlighted in red, while the refined keypoints from the fine stage are depicted in green. The visualized patches clearly illustrate the transformation of coarse keypoints from flat regions to peak points, demonstrating their effectiveness in enhancing the matching accuracy.}
\label{dynamic_refinement_results}
\end{figure}

\subsection{Dynamic Feature Refinement}
\label{method_dynamic_refinement}
Let $(P_x^A,P_y^B)$ be a pair of feature patches cropped at a coarse match $(x,y)$, where $x$ and $y$ are the center coordinates of $P_x^A$  and $P_y^B$, respectively. We propose a dynamic refinement network to estimate a refined match $(\Hat{x}, \Hat{y})$ at the finer stage. Existing methods \cite{sun2021loftr,chen2022aspanformer,tang2022quadtree} usually fix the first keypoint (i.e., $\Hat{x} = x$) and then employ an attention network to determine the location of the second keypoint $\Hat{y}$ with the highest similarity score. However, this approach can degrade the matching accuracy when the center location $x$ belongs to a flat region or a non-overlapping region of two patches. To address this issue, the refinement network estimates a score map for patch $P_x^A$, allowing it to detect a more reliable location as the keypoint of interest. 

First, we employed two MLP-Mixer blocks to transform the features of the two patches:
\begin{equation}
    \{\Hat{P}_x^A, \Hat{P}_y^B\} = \texttt{MLPMixer}\left(\{ P_x^A, P_y^B \}\right)
\end{equation}
The MLP-Mixer block consists of a channel-mixing block and a token-mixing block, each employing a multi-layer perceptron (MLP) and a LayerNorm layer. The channel-mixing block operates on each feature token independently, extracting new features by mixing channel information. On the other hand, the token-mixing block operates on each channel, facilitating communication between different spatial locations. Compared to a standard Transformer block, MLP-Mixer is more efficient \cite{tolstikhin2021mlp}. The detailed design of the MLP-Mixer block is illustrated in Fig. \ref{mlp_mixer_block}.

Next, we designed a detector network $\mathcal{D}$, also utilizing the MLP-Mixer blocks, to estimate a score map $S_x^A$ for feature patch $\Hat{P}_x^A$ as:
\begin{equation}
    S_x^A = \mathcal{D}(\Hat{P}_x^A), \quad (S_x^A \in \mathbb{R}^{N_p}, \Hat{P}_x^A \in \mathbb{R}^{N_p \times C})
\end{equation}
where $N_p$ is the number of features and $C$ is the number of channels. The score map $S_x^A$ was then converted to a probability heatmap $H_x^A$ using the softmax function with a small temperature $t$. Finally, we extracted the keypoint of interest $\Hat{x} \in \mathbb{R}^2$ and the corresponding feature descriptor $\Hat{f} \in \mathbb{R}^C$ using the expectation operation:
\begin{equation}
\label{eq:fine_kpts_estimation}
    \Hat{x} = \sum_{n=1}^{N_p} H_{x,n} G_n, \quad \Hat{f}_x^A = \sum_{n=1}^{N_p} H_{x,n}^A \Hat{P}_{x,n}^A
\end{equation}
Here, $G \in \mathbb{R}^{N_p \times 2}$ is a grid map indicating the spatial coordinates of features.

The entire network can be trained in a self-supervised manner, meaning it does not require ground-truth matching pairs. Moreover, it is observed that our network adaptively extracts keypoints by leveraging the information from both patches. As a result, the detected keypoint $\Hat{x}$ is highly matchable, thereby increasing the chance of finding the accurate matching keypoint $\Hat{y}$. The effectiveness of our approach is demonstrated in Fig. \ref{dynamic_refinement_results}, where the peak locations within the patches are detected as the refined results. 

Once we detected the interested keypoint for the first patch $P_x^A$, we estimated the matching keypoint in the second patch $P_y^B$. Given the detected feature $\Hat{f}_x^A$, we computed a heatmap $H_y^B$ to measure the similarity between $\Hat{f}_x^A$ and all feature points in $\Hat{P}_y^B$.
\begin{equation}
    H_y^B = \texttt{Softmax}\left( \langle \Hat{f}_x^A, \Hat{P}_y^B \rangle \right)
\end{equation}
Finally, the matching keypoint $\Hat{y}$ and the feature descriptor $\Hat{f}_y^B$ were determined using the expectation operator as in Eq. \ref{eq:fine_kpts_estimation}.

\subsection{Loss Function}
\label{method_loss_function}
The total loss is defined by the weighted sum over two loss components, coarse-level loss $\mathcal{L}_c$ and fine-level loss $\mathcal{L}_f$, as follows
\begin{equation}
    \mathcal{L} = \lambda_c \mathcal{L}_c + \lambda_f \mathcal{L}_f
\end{equation}
where $\lambda_c$ and $\lambda_f$ are constant weights for each loss term, and in this study, we set the weights as $\lambda_c = \lambda_f = 0.25$.

\begin{table*}[!t]
\caption{Evaluation of homography estimation on HPatches \cite{balntas2017hpatches}. We provide two experiments following the setups in LoFTR \cite{sun2021loftr} (left) and Patch2Pix \cite{zhou2021patch2pix} (right). \#M denotes the number of matches, and NN is the nearest neighbor algorithm}
\begin{minipage}[!t]{.37\textwidth}
    \centering
    \begin{tabular}{|m{8em}|ccc|c|}
    \hline
    \multirow{2}{*}{\textbf{Method}} & \multicolumn{3}{c|}{\textbf{AUC}} & \multirow{2}{*}{\textbf{\#M}} \\
    \cline{2-4}
    & 3px & 5px & 10px & \\
    \hline
    D2Net \cite{dusmanu2019d2} + NN & 23.2 & 35.9 & 53.6 & 0.2K \\
    R2D2 \cite{revaud2019r2d2} + NN & 50.6 & 63.9 & 76.8 & 0.5K \\
    DISK \cite{tyszkiewicz2020disk} + NN & 52.3 & 64.9 & 78.9 & 1.1K \\
    SP + SuperGlue & 53.9 & 68.4 & 81.7 & 0.6K \\
    Sparse-NCNet$^\star$ & 48.9 & 54.2 & 67.1 & 1.0K \\
    COTR$^\star$ \cite{jiang2021cotr} & 41.9 & 57.7 & 74.0 & 1.0K \\
    DRC-Net$^\star$ \cite{li2020dual} & 50.6 & 56.2 & 68.3 & 1.0K \\
    LoFTR$^\star$ \cite{sun2021loftr} & 65.9 & 75.6 & 84.6 & 1.0K \\
    PDC-Net+$^\star$ \cite{truong2023pdc} & 66.7 & 76.8 & 85.8 & 1.0K \\
    \hline
    TopicFM-fast$^\star$ & \textbf{70.5} & \textbf{79.7} & \textbf{87.6} & 1.0K \\
    TopicFM+$^\star$ & \textbf{70.9} & \textbf{80.2} & \textbf{88.3} & 1.0K \\
    \hline
  \end{tabular}
\end{minipage}
\hfill
\begin{minipage}[!t]{.63\textwidth}
    \centering
    \begin{tabular}{|m{12em}|ccc|c|}
    \hline
    \multirow{2}{*}{\textbf{Method}} & \multicolumn{3}{c|}{\textbf{Accuracy} ($\epsilon < 1/3/5$px)} & \multirow{2}{*}{\textbf{\#M}} \\
    \cline{2-4}
    & Overall & Illumination & Viewpoint & \\
    \hline
    ASLFeat \cite{luo2020aslfeat} + NN & 0.48/0.81/0.88 & 0.63/0.94/0.98 & 0.34/0.69/0.78 & 2.0K \\
    ASLFeat + SuperGlue \cite{sarlin2020superglue} & 0.49/0.83/0.89 & 0.57/0.92/0.98 & 0.41/0.72/0.81 & - \\
    ASLFeat + SGMNet \cite{chen2021learning} & 0.49/0.82/0.89 & 0.57/0.93/0.98 & 0.41/0.72/0.83 & - \\
    ASLFeat + ClusterGNN \cite{shi2022clustergnn} & 0.51/0.83/0.89 & 0.61/0.95/0.98 & 0.42/0.72/0.82	& - \\
    SP \cite{detone2018superpoint} + SuperGlue \cite{sarlin2020superglue} & 0.51/0.83/0.89 & 0.62/0.93/0.98 & 0.41/0.73/0.81 & 0.5K \\
    SP \cite{detone2018superpoint} + ClusterGNN \cite{shi2022clustergnn} & 0.52/0.84/0.90 & 0.61/0.93/0.98 & \textbf{0.44}/\textbf{0.74}/0.81 & - \\
    Sparse-NCNet$^\star$ \cite{rocco2020efficient} & 0.36/0.66/0.76 & 0.62/0.92/0.97 & 0.13/0.41/0.57 & 2.0K \\
    Patch2Pix$^\star$ \cite{zhou2021patch2pix} & 0.51/0.79/0.86 & 0.72/0.95/0.98 & 0.32/0.64/0.75 & 1.3K \\
    MatchFormer$^\star$ \cite{wang2022matchformer} & 0.55/0.81/0.87 & 0.75/0.95/0.98 & 0.37/0.68/0.78 & 4.8K \\
    \hline
    TopicFM-fast$^\star$ & \textbf{0.58}/\textbf{0.85}/\textbf{0.91} & \textbf{0.77}/\textbf{0.96}/\textbf{0.99} & 0.41/\textbf{0.74}/\textbf{0.83} & 5.1K \\
    TopicFM+$^\star$ & \textbf{0.59}/\textbf{0.85}/\textbf{0.92} & \textbf{0.76}/\textbf{0.96}/\textbf{0.99} & \textbf{0.44}/\textbf{0.74}/\textbf{0.85} & 4.9K \\
    \hline
  \end{tabular}
\end{minipage}
\label{table:hpatches}
\end{table*}

\subsubsection{Coarse-Level Loss} The coarse-level loss $(\mathcal{L}_c = \mathcal{L}_c^{feat} + \mathcal{L}_c^{topic})$ is comprised of the feature-matching loss $\mathcal{L}_c^{feat}$ and the topic-matching loss $\mathcal{L}_c^{topic}$, which are used to train both the network parameters and latent topics. The topic-matching loss was described in Section \ref{method_training_topics}. The feature-matching loss was computed from the matching probability matrix $P_c$ in Eq. \ref{eq:dual_softmax} by using cross-entropy function:
\begin{equation}
    \mathcal{L}_c^{feat} = - \sum_{(i,j) \in M_c^{gt}} \log(P_c(i, j))
\end{equation}

\subsubsection{Fine-Level Loss} We used the symmetric epipolar distance function \cite{hartley2003multiple} for the fine-level loss. This loss does not require explicit ground-truth matching pairs; therefore, it makes beneficial for training our self-detector network. Given a ground-truth fundamental matrix $F \in \mathbb{R}^{3 \times 3}$ and the matching coordinate pair $(\Hat{x}, \Hat{y})$ extracted from our method, the fine-level loss was defined as follows:
\begin{equation}
\label{eq:sym_epipolar_loss}
    \mathcal{L}_f = \frac{1}{M} \sum_{(\Hat{x}, \Hat{y})} \| \Hat{x}^T F \Hat{y} \|^2 \left( \frac{1}{\|F^T \Hat{x}\|_{0:2}^2} + \frac{1}{\|F \Hat{y}\|_{0:2}^2}\right)
\end{equation}
where $M$ represents the number of matching outputs. $\Hat{x}, \Hat{y}$ in Eq. \ref{eq:sym_epipolar_loss} above were already converted to homogeneous coordinates, and $\|v\|_{0:2}$  indicates that we only considered the first two elements of vector $v$.

\section{Experiments}
\label{exp}
\subsection{Experimental Setups}
\label{exp_setups}
\textbf{Training Details}. The proposed network was implemented in PyTorch, and the PyTorch-Lighting framework was employed for distributed training. The network was trained on the MegaDepth dataset \cite{li2018megadepth}, which comprises 196 scenes captured from diverse locations worldwide. The dataset underwent preprocessing to filter out low-quality scenes and divide each scene into sub-scenes based on covisible scores \cite{dusmanu2020multi}. For training, we utilized four GPUs, each with 11GB of memory. A subset of 368 sub-scenes was selected for training. In each training epoch, 50 image pairs were randomly sampled from each sub-scene, resulting in a total of 18,400 samples per epoch. To maintain consistent input sizes during training, the images were resized such that the highest dimension of each image was set to 800 pixels. 

The training process commenced with an initial learning rate of 0.001 and lasted for 40 epochs, completed in less than two days. In contrast to recent transformer-based models such as LoFTR \cite{sun2021loftr,chen2022aspanformer,wang2022matchformer}, which require approximately 24GB of memory per GPU and a day of training using 16 GPUs, our training approach is more efficient and well-suited for limited computational resources. During training, the number of topics $(K)$ was fixed at 100, and the confidence threshold $(\tau)$ for extracting coarse matches was set to 0.2.

\textbf{Evaluation}. We introduced two upgraded model variants: \textbf{TopicFM-fast} and \textbf{TopicFM+}. These models are extensions of our initial model, TopicFM \cite{khang2023topicfm}. Both variants incorporate dynamic feature refinement in the fine-level matching stage (Section \ref{method_dynamic_refinement}). The difference between the two variants lies in the implementation of the context merging step (Section \ref{method_context merging}) during the coarse-level matching stage. Specifically, TopicFM-fast utilizes an efficient context-merging strategy (Eq. \ref{eq:efficient_context_merging}),  while TopicFM+ estimates co-visible topics and employs in-topic self/cross-attention (Eq. \ref{eq:in_topic_feature_augmentation}).

We conducted a comprehensive evaluation of our models across multiple tasks and benchmarks, including: i) homography estimation, ii) relative pose estimation, iii) visual localization, and iv) image matching challenge. Additionally, we compared the efficiency of our models with state-of-the-art methods. It is important to note that our evaluation used models trained exclusively on the MegaDepth dataset, without any fine-tuning on other datasets. In the following sections, we will provide a detailed description of our evaluation process.

\begin{figure*}[!t]
\centering
\includegraphics[width=7.1in]{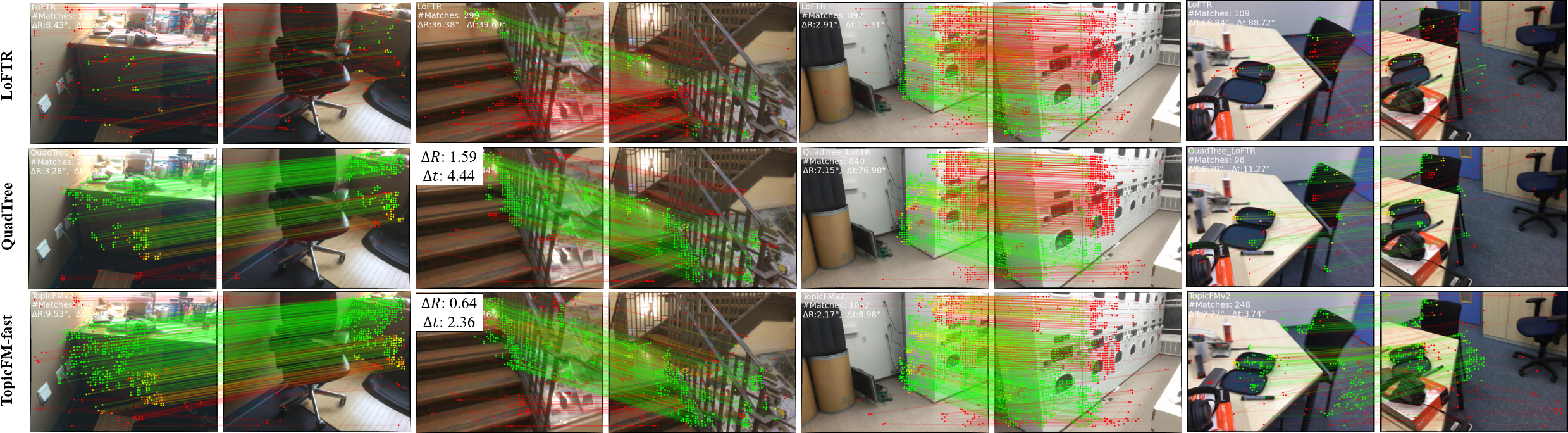}
\caption{Qualitative results of our method (TopicFM-fast) compared to LoFTR \cite{sun2021loftr} and QuadTree \cite{tang2022quadtree} on the ScanNet dataset. Our method exhibited superior performance compared to LoFTR and QuadTree, particularly in challenging conditions such as large relative viewpoints or untextured scenes.}
\label{fig_exp_scannet}
\end{figure*}

\subsection{Homography Estimation}
\label{exp_homography}
We utilized the HPatches dataset \cite{balntas2017hpatches} for evaluating this task. The HPatches dataset consists of 108 sequences, each containing one reference image and five target images. These image pairs were captured under varying illuminations and viewpoints, and ground-truth homography matrices between the reference and target images were provided. In our evaluation, we predicted feature correspondences for each image pair and estimated the homography matrix using OpenCV libraries. Previous image-matching methods have also conducted experiments on the HPatches dataset, although with different setups \cite{zhou2021patch2pix,sun2021loftr}.

We performed two versions of the evaluation, one based on LoFTR \cite{sun2021loftr} and the other on Patch2Pix \cite{zhou2021patch2pix}. In the first version, during the feature-matching step, we resized the image’s lowest dimension to 480. We selected an average of 1000 matches for each image pair to estimate the homography. In the second version, we resized the image’s highest dimension to 1024 and used all output correspondences for homography estimation. 

To assess the accuracy of the estimated homographies, we first warped the four corners of the first image to the second image based on the estimated and ground-truth homographies. We then computed the corner error $(\epsilon)$ between the two warped versions \cite{detone2018superpoint}. In the first evaluation, we measured the corner error using the AUC metric with thresholds of 3, 5, and 10 pixels \cite{sarlin2020superglue}. In the second evaluation, we calculated the percentage of correct estimates when the error $(\epsilon)$ was smaller than 1, 3, and 5 pixels.

The results are presented in Table \ref{table:hpatches}, which illustrates the significant accuracy improvements of the proposed models, TopicFM-fast and TopicFM+. As shown in Table \ref{table:hpatches} (left), both models outperformed the transformer-based method LoFTR \cite{sun2021loftr} and the recent dense matching method PDC-Net+ \cite{truong2023pdc} by a substantial margin. In Table \ref{table:hpatches} (right), our models achieved the best overall performance, surpassing both detector-based and detector-free methods. It is worth noting that, under the viewpoint-changing condition, only the SP+ClusterGNN baseline exhibited competitive performance. This method combines the robust keypoint detector SuperPoint (SP) \cite{detone2018superpoint} with the attention-based feature-matching model ClusterGNN \cite{shi2022clustergnn}. It is important to mention that both SP and ClusterGNN were trained independently on various datasets, whereas our models were trained end-to-end on MegaDepth, demonstrating the robustness of our proposed approach.

\begin{table}[t]
\caption{Evaluation of relative pose estimation on MegaDepth. $\star$ denotes detector-free methods.}
  \centering
  \begin{tabular}{|m{12em}|ccc|}
    \hline
    \multirow{2}{9em}{\textbf{Method}} & \multicolumn{3}{c|}{\textbf{Rel. Pose Est. (AUC)}} \\
    \cline{2-4}
    & \quad $5^o$ & \quad $10^o$ & \quad $20^o$ \\
    \hline
    SP \cite{detone2018superpoint} +  SuperGlue \cite{sarlin2020superglue} & 42.2 & 61.2 & 76.0 \\ 
    DRC-Net$^\star$ \cite{li2020dual} & 27.0	& 43.0	& 58.3 \\ 
    Patch2Pix$^\star$ \cite{zhou2021patch2pix} & 41.4 & 56.3 & 68.3 \\
    PDC-Net+(H)$^\star$ \cite{truong2023pdc} & 43.1	& 61.9 & 76.1 \\ 
    LoFTR$^\star$ \cite{sun2021loftr,shen2022semi} & 52.8 & 69.2 & 81.2 \\
    MatchFormer$^\star$ \cite{wang2022matchformer} & 52.9 & 69.7 & 82.0 \\
    QuadTree$\star$ \cite{tang2022quadtree} & 54.6 & 70.5 & 82.2 \\
    AspanFormer$^\star$ \cite{chen2022aspanformer} & 55.3 & 71.5 & \underline{83.1} \\
    \hline
    TopicFM-fast$^\star$ & \underline{56.2} & \underline{71.9} & 82.9 \\
    TopicFM+$^\star$ & \textbf{58.2} & \textbf{72.8} & \textbf{83.2} \\
    \hline
  \end{tabular} 
  \label{table:rel_pose_megadepth}
\end{table}

\begin{table}[t]
\caption{Relative pose estimation on Scannet. To make a fair comparison, we used models trained only on MegaDepth for the detector-free methods (denoted as $\star$).}
  \centering
  \begin{tabular}{|m{13em}|ccc|}
    \hline
    \multirow{2}{9em}{\textbf{Method}} & \multicolumn{3}{c|}{\textbf{Rel. Pose Est. (AUC)}} \\
    \cline{2-4}
    & \quad $5^o$ & \quad $10^o$ & \quad $20^o$ \\
    \hline
    ORB \cite{rublee2011orb} + GMS \cite{bian2017gms} & 5.21 & 13.65 & 25.36 \\ 
    D2-Net \cite{dusmanu2019d2} + NN & 5.25 & 14.53 & 27.96 \\
    ContextDesc \cite{luo2019contextdesc} + Ratio Test & 6.64 & 15.01 & 25.75 \\ 
    SP \cite{detone2018superpoint} + OANet \cite{zhang2019learning} & 11.76 & 26.90 & 43.85 \\
    SP \cite{detone2018superpoint} + SuperGlue \cite{sarlin2020superglue} & 16.16 & 33.81 & 51.84 \\ 
    DRC-Net$^\star$ \cite{li2020dual} & 7.69 & 17.93 & 30.49 \\
    MatchFormer$^\star$ \cite{wang2022matchformer} & 16.20 & 32.29 & 48.31 \\
    LoFTR$^\star$ \cite{shen2022semi} & 17.46 & 34.14 & 50.48 \\
    QuadTree$^\star$ \cite{tang2022quadtree} & 19.0 & 37.3 & 53.5 \\
    ASTR$^\star$ \cite{yu2023adaptive} & 19.4 & 37.6	& 54.4 \\
    \hline
    TopicFM-fast$^\star$ & 19.7 & 36.7 & 52.7 \\ 
    TopicFM+$^\star$ & \textbf{20.4} & \textbf{38.5} & \textbf{54.5} \\
    \hline
  \end{tabular} 
  
  \label{table:rel_pose_scannet}
\end{table}

\begin{figure*}[!t]
\centering
\includegraphics[width=7.1in]{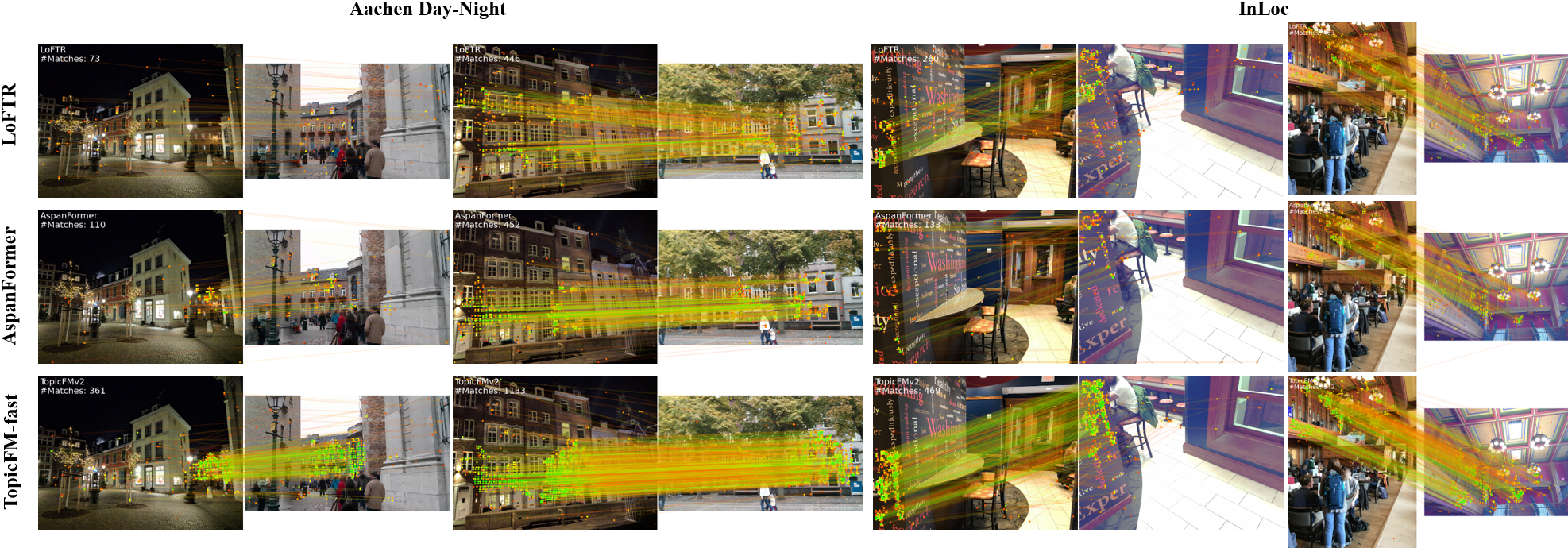}
\caption{Qualitative results on outdoor (Aachen Day-Night v1.1 \cite{zhang2021reference}) and indoor (InLoc \cite{taira2018inloc}) datasets. The matches are visualized using different colors to represent match confidence (green for high confidence, orange for low confidence), and matches with confidence below a threshold of 0.3 are filtered out. Our method produced a higher number of matches than LoFTR and AspanFormer in challenging conditions, such as day-night transitions, repetitive patterns, and illumination variations.}
\label{fig_exp_aachen_inloc}
\end{figure*}

\subsection{Relative Pose Estimation}
\label{exp_rel_pose_est}
Feature correspondences are commonly used for estimating the relative pose, which includes the rotation matrix $R$ and translation vector $T$, between two images. To evaluate the accuracy of relative pose estimation, we conducted evaluations on both outdoor (MegaDepth \cite{li2018megadepth}) and indoor (ScanNet \cite{dai2017scannet}) datasets. Each test set consisted of 1500 pairs of images. For the ScanNet dataset, we used an image resolution of $640\times480$, while for MegaDepth, we resized the highest dimension of the image to 1200. The relative pose for each image pair was estimated using OpenCV libraries.

Following the approach in \cite{sarlin2020superglue,sun2021loftr}, we measured the Area Under the Curve (AUC) of pose estimation error at thresholds of $\{5^o,10^o,20^o\}$. Tables \ref{table:rel_pose_megadepth} and \ref{table:rel_pose_scannet} present the results for the MegaDepth and ScanNet datasets, respectively. To ensure a fair comparison on ScanNet, we used models trained solely on MegaDepth for all detector-free methods. This helps to verify the models' generalization capacity. The proposed method, TopicFM+, outperformed the robust detector-based method SP \cite{detone2018superpoint} + SuperGlue \cite{sarlin2020superglue}, as well as recent Transformer-based detector-free methods, including LoFTR \cite{sun2021loftr}, MatchFormer \cite{wang2022matchformer}, QuadTree \cite{tang2022quadtree}, and ASTR \cite{yu2023adaptive}, as shown in Tables \ref{table:rel_pose_megadepth} and \ref{table:rel_pose_scannet}. Although our efficient version, TopicFM-fast, had lower performance compared to the robust version TopicFM+, it remained competitive with the Transformer-based baselines. Importantly, TopicFM+ demonstrated significant performance improvement on the ScanNet dataset, highlighting its strong generalization capabilities. On the MegaDepth dataset, our method achieved approximately an improvement of +3 AUC@$5^o$ compared to the second-best method, AspanFormer.

\subsection{Visual Localization}
\label{exp_vis_loc}
The visual localization task aims to estimate the absolute camera pose for an input query image based on a sparse 3D map of the scene. This task involves two main steps: finding a set of 2D-3D correspondences between the query image and the 3D map, and then using a perspective-n-points (PnP) algorithm to estimate the camera pose. The feature-matching model can be applied to the first step. In line with previous work \cite{zhou2021patch2pix,sun2021loftr}, we utilized the visual localization pipeline HLoc \cite{sarlin2019coarse} for evaluation. We integrated the matching model into this pipeline to assess its performance. HLoc also supports the construction of a 3D structure of the scene from a set of database images prior to performing query image localization. We evaluated both outdoor scenes (Aachen Day-Night v1.1 \cite{zhang2021reference}) and indoor scenes (InLoc \cite{taira2018inloc}). The estimated camera poses were submitted to the \href{https://www.visuallocalization.net/benchmark/}{benchmark website}, where the evaluation metrics were calculated.

\textbf{Aachen Day-Night v1.1}. This outdoor dataset comprises 6697 database images and 1015 query images, with 824 images captured during the day and 191 images taken at night. To generate image pairs, HLoc selected the top-20 nearest neighbors for each database image and the top-50 nearest neighbors for each query image. We resized the images to have the highest dimension of 1200 and predicted matches for each image pair. The matches were then aggregated across all image pairs to extract a unique set of matching keypoints for each image. This aggregation process involved keypoint quantization \cite{zhou2021patch2pix}. Subsequently, the image keypoints and matches were imported into the COLMAP database to perform SfM and absolute pose estimation. The entire pipeline typically required approximately one day to complete when executed on an NVIDIA Tesla V100 GPU with 32GB of memory.
The evaluation results are presented in Table \ref{table:vis_loc_aachen}, along with the results reported by other methods on the benchmark website. Our methods demonstrated the best performance across most evaluation metrics, outperforming the detector-based baseline SP \cite{detone2018superpoint} + SuperGlue \cite{sarlin2020superglue} and recent Transformer-based methods such as ASpanFormer \cite{chen2022aspanformer} and ASTR \cite{yu2023adaptive}. We observed that the detector-based method, SP+SuperGlue, achieved strong performance because SP and SuperGlue were trained on diverse datasets with various shapes and scenes, such as MS-COCO 2014 \cite{lin2014microsoft} (SP), synthetic shapes \cite{detone2018superpoint} (SP), and MegaDepth (SuperGlue). While ASpanFormer achieved good results, it required higher computational costs compared to LoFTR. In contrast, our method TopicFM-fast achieved both high performance and efficiency.

\textbf{InLoc}. This indoor dataset consists of 9972 database images and 329 query images. Since this dataset provides 3D pointcloud database, HLoc only generated query image pairs by selecting the top-40 nearest retrieval neighbors for each query image. During the feature-matching step, we set the highest dimension of the images to 1024. The entire InLoc experiment took approximately two hours on an NVIDIA Tesla V100 GPU with 32GB memory.
Table \ref{table:vis_loc_inloc} presents the evaluated results on InLoc. Remarkably, our method, TopicFM-fast, outperformed the other methods by a large margin. It also significantly surpassed our initial version of TopicFM \cite{khang2023topicfm}. This demonstrated the robustness of the proposed approach in this paper. 

\begin{table}[t]
\caption{Visual localization on Aachen Day-Night v1.1 \cite{zhang2021reference}. We reported the results using the HLoc pipeline \cite{sarlin2019coarse}.}
  \centering
  \begin{tabular}{|m{10em}|cc|c|}
    \hline
    \multirow{2}{*}{\textbf{Method}} & \textbf{Day} & \textbf{Night} & \multirow{2}{*}{\textbf{overall}} \\
    \cline{2-3}
    & \multicolumn{2}{c|}{(0.25m,$2^o$)/(0.5m,$5^o$)/(1.0m,$10^o$)} & \\
    \hline
    ISRF \cite{melekhov2020image} & 87.1/94.7/98.3 & 74.3/86.9/97.4 & 89.8 \\
    KAPTURE \cite{humenberger2020robust} + R2D2 \cite{revaud2019r2d2} + APGeM \cite{humenberger2020robust} & 90.0/\underline{96.2}/\textbf{99.5} & 72.3/86.4/97.9 & 90.4 \\
    SP + SuperGlue & 89.8/96.1/\underline{99.4} & 77.0/90.6/\textbf{100} & \underline{92.1} \\
    Patch2Pix \cite{zhou2021patch2pix} & 86.4/93.0/97.5 & 72.3/88.5/97.9 & 89.2 \\
    LoFTR \cite{sun2021loftr} & 88.7/95.6/99.0 & \underline{78.5}/90.6/99.0 & 91.9 \\
    ASpanFormer \cite{chen2022aspanformer} & 89.4/95.6/99.0 & 77.5/\underline{91.6}/\underline{99.5} & \underline{92.1} \\
    ASTR \cite{yu2023adaptive} & 89.9/95.6/99.2 & 76.4/\textbf{92.1}/\underline{99.5} & \underline{92.1} \\
    \hline
    TopicFM-fast & \textbf{90.4}/\textbf{96.4}/99.2 & \underline{78.5}/\underline{91.6}/\underline{99.5} & \textbf{92.6} \\
    TopicFM+ & \underline{90.2}/\underline{96.2}/99.2 & \textbf{79.1}/91.1/\underline{99.5} & \textbf{92.6} \\
    \hline
  \end{tabular}
  
  \label{table:vis_loc_aachen}
\end{table}
\begin{table}[t]
\caption{Visual localization on the InLoc dataset \cite{taira2018inloc} using HLoc. TopicFM-fast achieved the best overall performance.}
  \centering
  \begin{tabular}{|m{8em}|cc|c|}
    \hline
    \multirow{2}{*}{\textbf{Method}} & \textbf{DUC1} & \textbf{DUC2} & \multirow{2}{*}{\textbf{overall}} \\
    \cline{2-3}
    & \multicolumn{2}{c|}{(0.25m,$10^o$)/(0.5m,$10^o$)/(1.0m,$10^o$)} & \\
    \hline
    ISRF \cite{melekhov2020image} & 39.4/58.1/70.2 & 41.2/61.1/69.5 & 56.6 \\ 
    KAPTURE \cite{humenberger2020robust} + R2D2 \cite{revaud2019r2d2} & 41.4/60.1/73.7 & 47.3/67.2/73.3 & 60.5 \\
    SP + SuperGlue & 49.0/68.7/80.8 & 53.4/\textbf{77.1}/82.4 & 68.6 \\
    Patch2Pix \cite{zhou2021patch2pix} & 44.4/66.7/78.3 & 49.6/64.9/72.5 & 62.7 \\
    LoFTR \cite{sun2021loftr} & 47.5/72.2/84.8 & 54.2/74.8/\textbf{85.5} & 69.8 \\
    MatchFormer \cite{wang2022matchformer} & 46.5/73.2/85.9 & \underline{55.7}/71.8/81.7 & 69.1 \\
    ASpanFormer \cite{chen2022aspanformer} & 51.5/73.7/\underline{86.4} & 55.0/74.0/81.7 & 70.4 \\
    ASTR \cite{yu2023adaptive} & \textbf{53.0}/73.7/\textbf{87.4} & 52.7/\underline{76.3}/84.0 & \underline{71.2} \\
    \hline
    TopicFM \cite{khang2023topicfm} & 52.0/74.7/\textbf{87.4} & 53.4/74.8/83.2 & 70.9 \\
    TopicFM-fast & \textbf{53.0}/\textbf{76.3}/\underline{86.4} & \textbf{57.3}/74.8/\underline{84.7} & \textbf{72.1} \\

    \hline
  \end{tabular}
  
  \label{table:vis_loc_inloc}
\end{table}

\begin{table}[t]
\caption{Relative Pose Estimation of Image Matching Challenge 2022. The baseline results are reported from Kaggle (\href{https://www.kaggle.com/competitions/image-matching-challenge-2022/discussion/328805}{Link})}
  \centering
  \begin{tabular}{|m{12em}|cc|}
    \hline
    \multirow{2}{*}{\textbf{Method}} & \multicolumn{2}{c|}{\textbf{Relative Pose Estimation (mAA)}} \\
    \cline{2-3}
    & Private Score & Public Score \\
    \hline
    SP \cite{detone2018superpoint} + SuperGlue \cite{sarlin2020superglue} & 0.724 & 0.728 \\
    LoFTR \cite{sun2021loftr} & 0.786 & 0.777 \\
    MatchFormer \cite{wang2022matchformer} & 0.783 & 0.774  \\
    QuadTree \cite{tang2022quadtree} & 0.817 & 0.812 \\
    \hline
    TopicFM-fast & \textbf{0.823} & \textbf{0.821} \\
    \hline
  \end{tabular}
  
  \label{table:imc2022}
\end{table}

\begin{figure}[!t]
\centering
\includegraphics[width=3in]{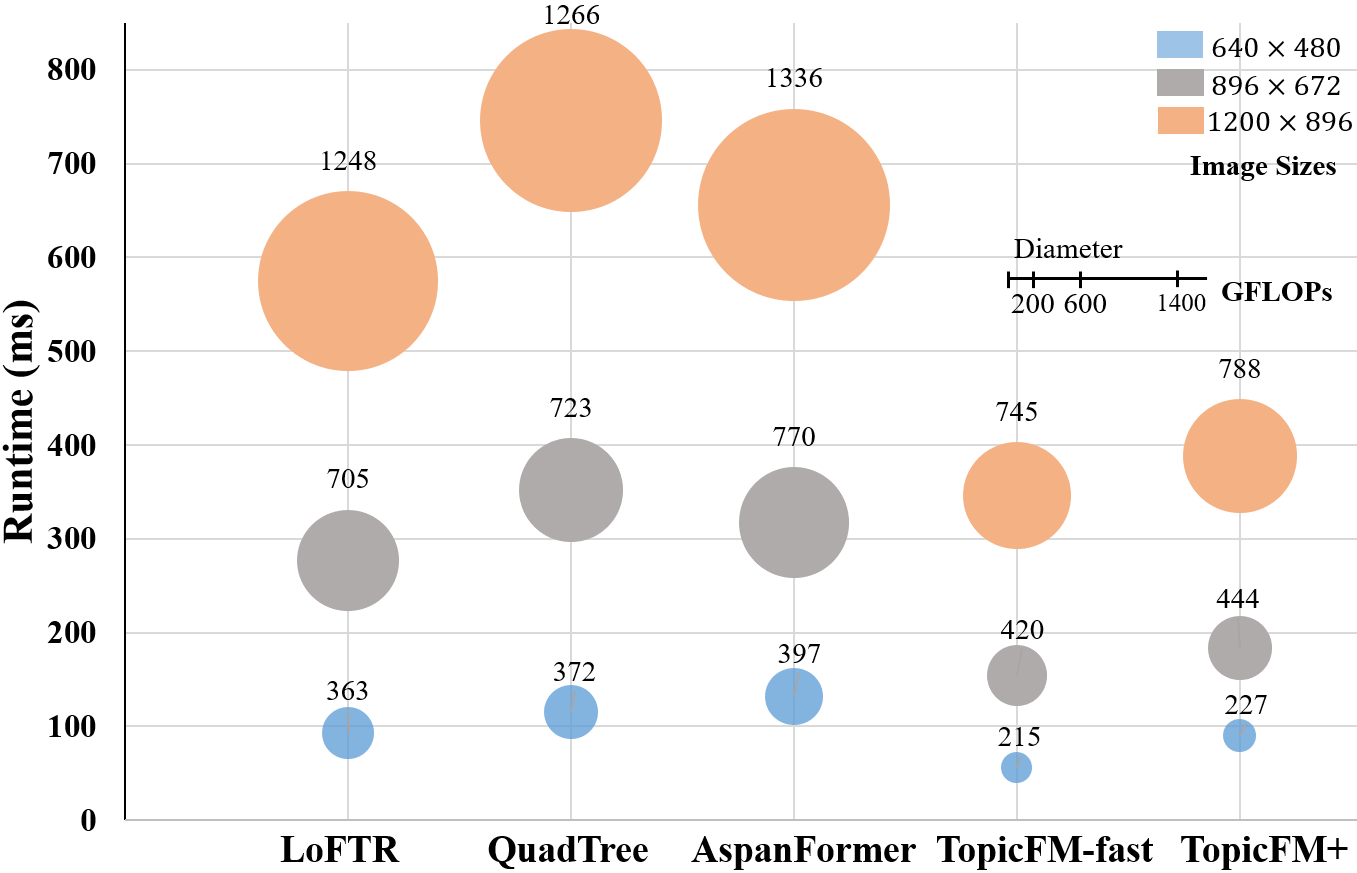}
\caption{Efficiency comparison between our models and the other transformer-based models. The diameter of each circle represents the computational cost measured in GFLOPs.}
\label{fig_exp_efficiency}
\end{figure}

\subsection{Image Matching Challenge}
\label{exp_imc22}
Image Matching Challenge 2022 (\href{https://www.kaggle.com/competitions/image-matching-challenge-2022}{IMC22}) was hosted on Kaggle. We submitted the source code to the benchmark website and received a mean Average Accuracy (mAA) score for camera pose estimation on a hidden dataset. For the experimental setups, we resized the images to have the highest dimension of 1472. After the image-matching step, we used OpenCV libraries and utilized a RANSAC threshold of 0.15 pixels to estimate a fundamental matrix for each image pair. These estimated fundamental matrices underwent post-processing and evaluation to obtain the final output scores.

Table \ref{table:imc2022} shows the results of the proposed method compared to other image-matching methods. TopicFM-fast achieved better accuracy than the baselines. Here, we did not report the highest-ranking submissions because those submissions employed ensemble techniques that combine the predictions of multiple image-matching models to produce the final prediction.

\subsection{Efficiency Evaluation}
\label{exp_efficiency}
We compared our method with recent Transformer-based methods, including LoFTR \cite{sun2021loftr}, QuadTree \cite{tang2022quadtree}, and AspanFormer \cite{chen2022aspanformer}. These methods utilized a ResNet-based network for multi-scale feature extraction and incorporated attention layers for coarse-level and fine-level matching. To evaluate efficiency, we measured GFLOPs (1 GFLOPs = $10^6$ FLOPs) and runtime (in ms) of our method and the comparison methods. We calculated the average values of these metrics using 1500 image pairs from the ScanNet dataset. The measurements were performed on a workstation with an Intel Xeon CPU (48 cores), 252 GB of RAM, and a Tesla V100 GPU with 32 GB of memory. Fig. \ref{fig_exp_efficiency} shows the experimental results at various image resolutions. As shown in Fig. \ref{fig_exp_efficiency}, TopicFM-fast outperformed TopicFM+ in terms of computational cost and runtime due to the efficient context merging module described in Section \ref{method_context merging}. Both models exhibited significantly reduced computational costs and runtimes as the image resolution increased. Compared to the state-of-the-art method AspanFormer, TopicFM-fast achieved approximately a 50\% reduction in runtime and GFLOPs, while demonstrating superior image-matching performance across most benchmarks. 
\begin{figure*}[t]
\centering
\includegraphics[width=7in]{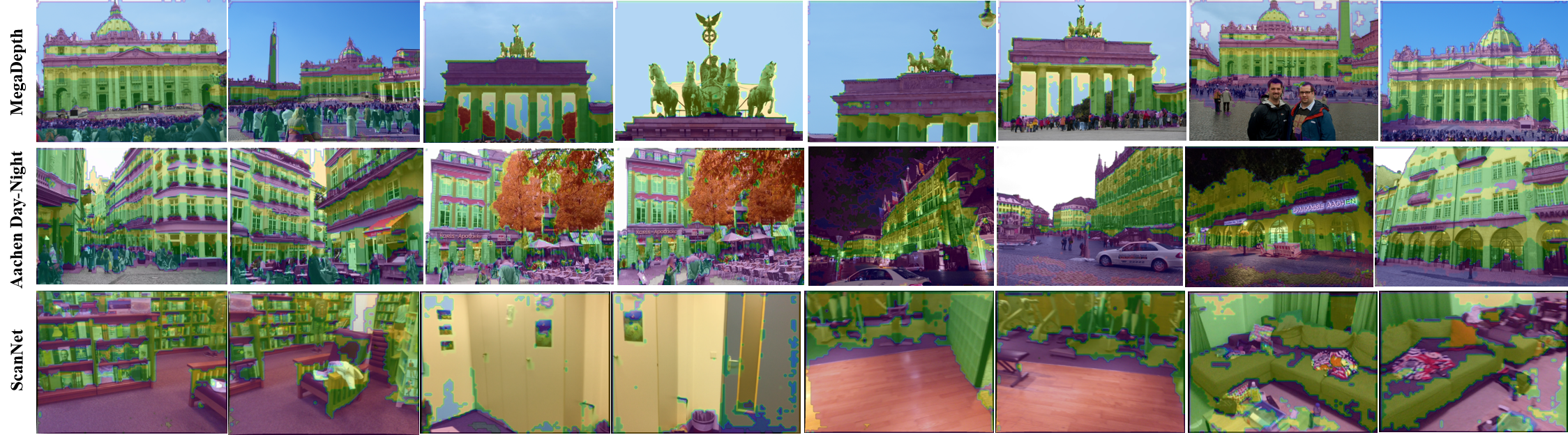}
\caption{Illustration of the topics estimated in the topic-assisted feature matching module (Section \ref{method_topicfm}). We observe that various types of spatial structures and objects across images and datasets are represented by different topics. These topics provide sufficient context information and facilitate the recognition of co-visible regions, effectively supporting the matching process.}
\label{fig_topic_visualization}
\end{figure*}
\subsection{Topic Visualization}
\label{exp_topic_viz}
The latent topics offer interpretability to the image-matching results, providing insights into the contextual information used for feature learning and indicating the presence of overlapping regions between images. To visualize the topics within each image, we selected the topic label with the highest probability based on the topic distribution $\theta$ of each feature. Fig. \ref{fig_topic_visualization} illustrates the results, where each topic in the image is represented by a specific color. As shown in Fig. \ref{fig_topic_visualization}, distinct spatial structures or objects consistently correspond to the same topic in each image. For example, in the first two image pairs of MegaDepth and Aachen, the topic "human" is highlighted in green, "tree" in orange, and "ground" in purple. Different spatial structures of a building, such as roofs, windows, and pillars, are assigned to different topics. This consistent pattern across images from both the MegaDepth and Aachen Day-Night datasets demonstrates the effectiveness of our topic inference module. Furthermore, as depicted in the last two image pairs of the MegaDepth and Aachen datasets, our method can identify overlapping structures by considering the co-visible topics (indicated with color) while disregarding non-overlapping regions (marked without color). Remarkably, despite being trained on the outdoor MegaDepth dataset, our topic-assisted feature-matching approach shows good generalization on the indoor ScanNet dataset, as evident in the last row of Fig. \ref{fig_topic_visualization}.

\begin{table}[t]
\caption{Analysis of the proposed modules. We evaluated the effectiveness of each module by comparing different model variants. We trained these models at the image resolution of $800 \times 800$ and tested at the resolution of $864 \times 864$.}
  \centering
  \begin{tabular}{|c|c|c|c|c|c|}
    \hline
     \multirow{2}{*}{Modules} & \multirow{2}{*}{FPN} & \ref{method_topicfm} & \ref{method_topicfm} & \multirow{2}{*}{\ref{method_dynamic_refinement}} & AUC \\
     & & (Eq. \ref{eq:in_topic_feature_augmentation}) & (Eq. \ref{eq:efficient_context_merging}) & & ($5^o/10^o/20^o$) \\
    \hline
    Model-A & $\checkmark$ & $\checkmark$ & & & 52.3 / 68.6 / 80.9 \\ 
    Model-B & $\checkmark$ & & $\checkmark$ & & 52.2 / 68.7 / 80.7 \\ 
    Model-C & $\checkmark$ & & $\checkmark$ & $\checkmark$ & 53.2 / 69.4 / 81.5 \\ 
    Model-D & $\checkmark$ & $\checkmark$ & & $\checkmark$ & 53.5 / 70.1 / 81.8 \\ 
    \hline
    GFLOPs & 462 & 73 & 49 & 10 & \\
    \hline
  \end{tabular}
  \label{table:ablation_study_modules}
\end{table}

\begin{table}[t]
\caption{Influence of the co-visible topics ($K_{co}$). We tested the TopicFM+ model using the MegaDepth dataset, in which the highest dimension of each image was resized to 1200.}
  \centering
  \begin{tabular}{|c|ccc|c|}
    \hline
    \multirow{2}{*}{$K_{co}$} & \multicolumn{3}{c|}{\textbf{Relative Pose Estimation}} & \multirow{2}{*}{\textbf{Runtime (ms)}} \\
    \cline{2-4}
    & AUC@$5^o$ & AUC@$10^o$ & AUC@$20^o$ & \\
    \hline
    2 & 47.1 & 61.3 & 71.9 & 308 \\ 
    4 & 54.7 & 70.3 & 81.4 & 332 \\
    6 & 56.8 & 72.0 & 82.9 & 346 \\
    8 & \textbf{58.2} & \textbf{72.8} & \textbf{83.2} & 358 \\
    10 & 58.0 & 72.5 & 83.0 & 362 \\
    12 & 57.9 & 72.4 & 83.0 & 362 \\
    \hline
  \end{tabular}
  \label{table:ablation_study_covisible}
\end{table}

\subsection{Ablation Study}
\label{ablation_study}
In this section, we conducted experiments on the MegaDepth dataset to analyze the effectiveness and computational efficiency of the key components in our method. The proposed method consists of three modules: (i) the feature initialization FPN, (ii) the coarse-level matching module using TopicFM (Section \ref{method_topicfm}), and (iii) the fine-level matching module using dynamic feature refinement (Section \ref{method_dynamic_refinement}). For the coarse-level matching step, we investigated two versions of context merging: in-topic feature augmentation (Eq. \ref{eq:in_topic_feature_augmentation}) and efficient context merging (Eq. \ref{eq:efficient_context_merging}).

\textbf{Efficiency Details}. We measured the computational cost for each module at an image resolution of 864×864. As shown in Table \ref{table:ablation_study_modules}, the feature extraction step incurred the highest computational cost. This CNN-based network, commonly employed in other methods, is crucial for initializing robust features for subsequent steps. Furthermore, the proposed efficient context merging technique (Eq. \ref{eq:efficient_context_merging}) demonstrates higher efficiency compared to the in-topic feature augmentation approach (Eq. \ref{eq:in_topic_feature_augmentation}).

\textbf{Effectiveness of Proposed Modules}. To investigate the impact of each module, we conducted experiments comparing the performance gains achieved by activating each module sequentially. We tested the following models:
\begin{itemize}
    \item Model-A represents our original work, TopicFM \cite{khang2023topicfm}. It used the feature extraction network FPN and employed in-topic feature augmentation (Eq. \ref{eq:in_topic_feature_augmentation}) for the coarse-level matching stage. For the fine-level matching stage, Model-A adopted the fixed center-keypoint approach, similar to LoFTR \cite{sun2021loftr}.
    \item Model-B improved the efficiency of Model-A by incorporating the efficient context merging technique (Eq. \ref{eq:efficient_context_merging}) in the coarse-level matching step.
    \item Model-C built upon Model-B by further applying dynamic feature refinement in the fine-level matching step. Model-C represents the proposed TopicFM-fast in this paper.
    \item Model-D represents TopicFM+ that employed in-topic feature augmentation and dynamic refinement.
\end{itemize}

Based on Table \ref{table:ablation_study_modules}, we can make two key observations. Firstly, there is a tradeoff between matching accuracy and computational cost when transitioning from in-topic feature augmentation (Eq. \ref{eq:in_topic_feature_augmentation}) to efficient context merging (Eq. \ref{eq:efficient_context_merging}) in the coarse-level matching step. The efficient context merging module enhances computational efficiency (from 73 GFLOPs to 49 GFLOPs) at the expense of a slight decrease in accuracy (Model-A to Model-B, Model-D to Model-C). Secondly, the dynamic feature refinement module significantly improves accuracy (Model-A to Model-D, Model-B to Model-C). In summary, we suggested several options to improve image-matching accuracy or efficiency. Based on the evaluations conducted in previous sections, we conclude that TopicFM-fast (Model-C) balances accuracy and computational cost, making it a valuable advancement.

\textbf{Covisible Topic Selection}. The proposed method utilizes the estimation of covisible probabilities of topics (as described in Section \ref{method_context merging}) to select important topics for robust feature learning. To assess the impact of the number of covisible topics, denoted as $K_{co}$, on image-matching performance, we conducted evaluations on MegaDepth using AUC metrics. Table \ref{table:ablation_study_covisible} presents the results of relative pose estimation and corresponding runtimes for each $K_{co} \in \{2,4,6,8,10,12\}$. It is observed that the matching accuracy gradually improved as $K_{co}$ increased. However, there was a slight decrease in AUCs when $K_{co}$ exceeded 8, and the reduction stopped after reaching 10. This indicates that TopicFM effectively covered all covisible regions in the image pair. In conclusion, it is recommended to set the number of covisible topics between 6 and 10 during testing to achieve high image-matching accuracy.

\section{CONCLUSION}
\label{conclusion}
We have presented a novel method for solving the image-matching problem. Our approach leverages latent topics within each image to effectively support the feature-matching process. Our experiments demonstrate that the topic-assisted feature-matching approach achieves high accuracy while maintaining low computational costs. Additionally, we introduced a dynamic feature refinement technique to enhance pixel-level accuracy. As a result, the proposed architecture is robust and significantly more efficient than state-of-the-art methods. We are confident that our method will make a substantial contribution to real-time applications.

\section*{Acknowledgments}
This work was supported by the Industrial Strategic Technology Development Program (No. 20007058, Development of safe and comfortable human augmentation hybrid robot suit) funded by the Ministry of Trade, Industry, \& Energy (MOTIE, Korea).



 

\bibliographystyle{IEEEtran}
\bibliography{refs}

\begin{thebibliography}{10}
\providecommand{\url}[1]{#1}
\csname url@samestyle\endcsname
\providecommand{\newblock}{\relax}
\providecommand{\bibinfo}[2]{#2}
\providecommand{\BIBentrySTDinterwordspacing}{\spaceskip=0pt\relax}
\providecommand{\BIBentryALTinterwordstretchfactor}{4}
\providecommand{\BIBentryALTinterwordspacing}{\spaceskip=\fontdimen2\font plus
\BIBentryALTinterwordstretchfactor\fontdimen3\font minus
  \fontdimen4\font\relax}
\providecommand{\BIBforeignlanguage}[2]{{%
\expandafter\ifx\csname l@#1\endcsname\relax
\typeout{** WARNING: IEEEtran.bst: No hyphenation pattern has been}%
\typeout{** loaded for the language `#1'. Using the pattern for}%
\typeout{** the default language instead.}%
\else
\language=\csname l@#1\endcsname
\fi
#2}}
\providecommand{\BIBdecl}{\relax}
\BIBdecl

\bibitem{schonberger2016structure}
J.~L. Schonberger and J.-M. Frahm, ``Structure-from-motion revisited,'' in
  \emph{Proceedings of the IEEE conference on computer vision and pattern
  recognition}, 2016, pp. 4104--4113.

\bibitem{zhu2018very}
S.~Zhu, R.~Zhang, L.~Zhou, T.~Shen, T.~Fang, P.~Tan, and L.~Quan, ``Very
  large-scale global sfm by distributed motion averaging,'' in
  \emph{Proceedings of the IEEE conference on computer vision and pattern
  recognition}, 2018, pp. 4568--4577.

\bibitem{mur2015orb}
R.~Mur-Artal, J.~M.~M. Montiel, and J.~D. Tardos, ``Orb-slam: a versatile and
  accurate monocular slam system,'' \emph{IEEE transactions on robotics},
  vol.~31, no.~5, pp. 1147--1163, 2015.

\bibitem{campos2021orb}
C.~Campos, R.~Elvira, J.~J.~G. Rodr{\'\i}guez, J.~M. Montiel, and J.~D.
  Tard{\'o}s, ``Orb-slam3: An accurate open-source library for visual,
  visual--inertial, and multimap slam,'' \emph{IEEE Transactions on Robotics},
  vol.~37, no.~6, pp. 1874--1890, 2021.

\bibitem{tomasi1991tracking}
C.~Tomasi and T.~K. Detection, ``Tracking of point features,'' \emph{Int. J.
  Comput. Vis}, vol.~9, pp. 137--154, 1991.

\bibitem{shi1994good}
J.~Shi \emph{et~al.}, ``Good features to track,'' in \emph{1994 Proceedings of
  IEEE conference on computer vision and pattern recognition}.\hskip 1em plus
  0.5em minus 0.4em\relax IEEE, 1994, pp. 593--600.

\bibitem{sarlin2019coarse}
P.-E. Sarlin, C.~Cadena, R.~Siegwart, and M.~Dymczyk, ``From coarse to fine:
  Robust hierarchical localization at large scale,'' in \emph{Proceedings of
  the IEEE/CVF Conference on Computer Vision and Pattern Recognition}, 2019,
  pp. 12\,716--12\,725.

\bibitem{lynen2020large}
S.~Lynen, B.~Zeisl, D.~Aiger, M.~Bosse, J.~Hesch, M.~Pollefeys, R.~Siegwart,
  and T.~Sattler, ``Large-scale, real-time visual--inertial localization
  revisited,'' \emph{The International Journal of Robotics Research}, vol.~39,
  no.~9, pp. 1061--1084, 2020.

\bibitem{lowe2004distinctive}
D.~G. Lowe, ``Distinctive image features from scale-invariant keypoints,''
  \emph{International journal of computer vision}, vol.~60, no.~2, pp. 91--110,
  2004.

\bibitem{bay2008speeded}
H.~Bay, A.~Ess, T.~Tuytelaars, and L.~Van~Gool, ``Speeded-up robust features
  (surf),'' \emph{Computer vision and image understanding}, vol. 110, no.~3,
  pp. 346--359, 2008.

\bibitem{calonder2011brief}
M.~Calonder, V.~Lepetit, M.~Ozuysal, T.~Trzcinski, C.~Strecha, and P.~Fua,
  ``Brief: Computing a local binary descriptor very fast,'' \emph{IEEE
  transactions on pattern analysis and machine intelligence}, vol.~34, no.~7,
  pp. 1281--1298, 2011.

\bibitem{sattler2012improving}
T.~Sattler, B.~Leibe, and L.~Kobbelt, ``Improving image-based localization by
  active correspondence search,'' in \emph{European conference on computer
  vision}.\hskip 1em plus 0.5em minus 0.4em\relax Springer, 2012, pp. 752--765.

\bibitem{jin2021image}
Y.~Jin, D.~Mishkin, A.~Mishchuk, J.~Matas, P.~Fua, K.~M. Yi, and E.~Trulls,
  ``Image matching across wide baselines: From paper to practice,''
  \emph{International Journal of Computer Vision}, vol. 129, no.~2, pp.
  517--547, 2021.

\bibitem{yi2016lift}
K.~M. Yi, E.~Trulls, V.~Lepetit, and P.~Fua, ``Lift: Learned invariant feature
  transform,'' in \emph{European conference on computer vision}.\hskip 1em plus
  0.5em minus 0.4em\relax Springer, 2016, pp. 467--483.

\bibitem{ono2018lf}
Y.~Ono, E.~Trulls, P.~Fua, and K.~M. Yi, ``Lf-net: Learning local features from
  images,'' \emph{Advances in neural information processing systems}, vol.~31,
  2018.

\bibitem{detone2018superpoint}
D.~DeTone, T.~Malisiewicz, and A.~Rabinovich, ``Superpoint: Self-supervised
  interest point detection and description,'' in \emph{Proceedings of the IEEE
  conference on computer vision and pattern recognition workshops}, 2018, pp.
  224--236.

\bibitem{li2020dual}
X.~Li, K.~Han, S.~Li, and V.~Prisacariu, ``Dual-resolution correspondence
  networks,'' \emph{Advances in Neural Information Processing Systems},
  vol.~33, pp. 17\,346--17\,357, 2020.

\bibitem{rocco2020efficient}
I.~Rocco, R.~Arandjelovi{\'c}, and J.~Sivic, ``Efficient neighbourhood
  consensus networks via submanifold sparse convolutions,'' in \emph{European
  conference on computer vision}.\hskip 1em plus 0.5em minus 0.4em\relax
  Springer, 2020, pp. 605--621.

\bibitem{rocco2018neighbourhood}
I.~Rocco, M.~Cimpoi, R.~Arandjelovi{\'c}, A.~Torii, T.~Pajdla, and J.~Sivic,
  ``Neighbourhood consensus networks,'' \emph{Advances in neural information
  processing systems}, vol.~31, 2018.

\bibitem{sun2021loftr}
J.~Sun, Z.~Shen, Y.~Wang, H.~Bao, and X.~Zhou, ``Loftr: Detector-free local
  feature matching with transformers,'' in \emph{Proceedings of the IEEE/CVF
  conference on computer vision and pattern recognition}, 2021, pp. 8922--8931.

\bibitem{wang2022matchformer}
Q.~Wang, J.~Zhang, K.~Yang, K.~Peng, and R.~Stiefelhagen, ``Matchformer:
  Interleaving attention in transformers for feature matching,'' in
  \emph{Proceedings of the Asian Conference on Computer Vision}, 2022, pp.
  2746--2762.

\bibitem{jiang2021cotr}
W.~Jiang, E.~Trulls, J.~Hosang, A.~Tagliasacchi, and K.~M. Yi, ``Cotr:
  Correspondence transformer for matching across images,'' in \emph{Proceedings
  of the IEEE/CVF International Conference on Computer Vision}, 2021, pp.
  6207--6217.

\bibitem{chen2022aspanformer}
H.~Chen, Z.~Luo, L.~Zhou, Y.~Tian, M.~Zhen, T.~Fang, D.~McKinnon, Y.~Tsin, and
  L.~Quan, ``Aspanformer: Detector-free image matching with adaptive span
  transformer,'' in \emph{Computer Vision--ECCV 2022: 17th European Conference,
  Tel Aviv, Israel, October 23--27, 2022, Proceedings, Part XXXII}.\hskip 1em
  plus 0.5em minus 0.4em\relax Springer, 2022, pp. 20--36.

\bibitem{dosovitskiy2020image}
A.~Dosovitskiy, L.~Beyer, A.~Kolesnikov, D.~Weissenborn, X.~Zhai,
  T.~Unterthiner, M.~Dehghani, M.~Minderer, G.~Heigold, S.~Gelly \emph{et~al.},
  ``An image is worth 16x16 words: Transformers for image recognition at
  scale,'' \emph{arXiv preprint arXiv:2010.11929}, 2020.

\bibitem{dai2023oamatcher}
K.~Dai, T.~Xie, K.~Wang, Z.~Jiang, R.~Li, and L.~Zhao, ``Oamatcher: An
  overlapping areas-based network for accurate local feature matching,''
  \emph{arXiv preprint arXiv:2302.05846}, 2023.

\bibitem{zhou2021patch2pix}
Q.~Zhou, T.~Sattler, and L.~Leal-Taixe, ``Patch2pix: Epipolar-guided
  pixel-level correspondences,'' in \emph{Proceedings of the IEEE/CVF
  conference on computer vision and pattern recognition}, 2021, pp. 4669--4678.

\bibitem{li2018megadepth}
Z.~Li and N.~Snavely, ``Megadepth: Learning single-view depth prediction from
  internet photos,'' in \emph{Proceedings of the IEEE conference on computer
  vision and pattern recognition}, 2018, pp. 2041--2050.

\bibitem{blei2003latent}
D.~M. Blei, A.~Y. Ng, and M.~I. Jordan, ``Latent dirichlet allocation,''
  \emph{Journal of machine Learning research}, vol.~3, no. Jan, pp. 993--1022,
  2003.

\bibitem{yan2013biterm}
X.~Yan, J.~Guo, Y.~Lan, and X.~Cheng, ``A biterm topic model for short texts,''
  in \emph{Proceedings of the 22nd international conference on World Wide Web},
  2013, pp. 1445--1456.

\bibitem{hartley2003multiple}
R.~Hartley and A.~Zisserman, \emph{Multiple view geometry in computer
  vision}.\hskip 1em plus 0.5em minus 0.4em\relax Cambridge university press,
  2003.

\bibitem{tolstikhin2021mlp}
I.~O. Tolstikhin, N.~Houlsby, A.~Kolesnikov, L.~Beyer, X.~Zhai, T.~Unterthiner,
  J.~Yung, A.~Steiner, D.~Keysers, J.~Uszkoreit \emph{et~al.}, ``Mlp-mixer: An
  all-mlp architecture for vision,'' \emph{Advances in neural information
  processing systems}, vol.~34, pp. 24\,261--24\,272, 2021.

\bibitem{tang2022quadtree}
S.~Tang, J.~Zhang, S.~Zhu, and P.~Tan, ``Quadtree attention for vision
  transformers,'' \emph{arXiv preprint arXiv:2201.02767}, 2022.

\bibitem{khang2023topicfm}
\BIBentryALTinterwordspacing
K.~Truong~Giang, S.~Song, and S.~Jo, ``Topicfm: Robust and interpretable
  topic-assisted feature matching,'' \emph{Proceedings of the AAAI Conference
  on Artificial Intelligence}, vol.~37, no.~2, pp. 2447--2455, Jun. 2023.
  [Online]. Available:
  \url{https://ojs.aaai.org/index.php/AAAI/article/view/25341}
\BIBentrySTDinterwordspacing

\bibitem{rosten2006machine}
E.~Rosten and T.~Drummond, ``Machine learning for high-speed corner
  detection,'' in \emph{European conference on computer vision}.\hskip 1em plus
  0.5em minus 0.4em\relax Springer, 2006, pp. 430--443.

\bibitem{noh2017large}
H.~Noh, A.~Araujo, J.~Sim, T.~Weyand, and B.~Han, ``Large-scale image retrieval
  with attentive deep local features,'' in \emph{Proceedings of the IEEE
  international conference on computer vision}, 2017, pp. 3456--3465.

\bibitem{dusmanu2019d2}
M.~Dusmanu, I.~Rocco, T.~Pajdla, M.~Pollefeys, J.~Sivic, A.~Torii, and
  T.~Sattler, ``D2-net: A trainable cnn for joint description and detection of
  local features,'' in \emph{Proceedings of the ieee/cvf conference on computer
  vision and pattern recognition}, 2019, pp. 8092--8101.

\bibitem{revaud2019r2d2}
J.~Revaud, P.~Weinzaepfel, C.~De~Souza, N.~Pion, G.~Csurka, Y.~Cabon, and
  M.~Humenberger, ``R2d2: repeatable and reliable detector and descriptor,''
  \emph{arXiv preprint arXiv:1906.06195}, 2019.

\bibitem{bhowmik2020reinforced}
A.~Bhowmik, S.~Gumhold, C.~Rother, and E.~Brachmann, ``Reinforced feature
  points: Optimizing feature detection and description for a high-level task,''
  in \emph{Proceedings of the IEEE/CVF conference on computer vision and
  pattern recognition}, 2020, pp. 4948--4957.

\bibitem{tyszkiewicz2020disk}
M.~Tyszkiewicz, P.~Fua, and E.~Trulls, ``Disk: Learning local features with
  policy gradient,'' \emph{Advances in Neural Information Processing Systems},
  vol.~33, pp. 14\,254--14\,265, 2020.

\bibitem{muja2014scalable}
M.~Muja and D.~G. Lowe, ``Scalable nearest neighbor algorithms for high
  dimensional data,'' \emph{IEEE transactions on pattern analysis and machine
  intelligence}, vol.~36, no.~11, pp. 2227--2240, 2014.

\bibitem{luo2019contextdesc}
Z.~Luo, T.~Shen, L.~Zhou, J.~Zhang, Y.~Yao, S.~Li, T.~Fang, and L.~Quan,
  ``Contextdesc: Local descriptor augmentation with cross-modality context,''
  in \emph{Proceedings of the IEEE/CVF conference on computer vision and
  pattern recognition}, 2019, pp. 2527--2536.

\bibitem{luo2020aslfeat}
Z.~Luo, L.~Zhou, X.~Bai, H.~Chen, J.~Zhang, Y.~Yao, S.~Li, T.~Fang, and
  L.~Quan, ``Aslfeat: Learning local features of accurate shape and
  localization,'' in \emph{Proceedings of the IEEE/CVF conference on computer
  vision and pattern recognition}, 2020, pp. 6589--6598.

\bibitem{sarlin2020superglue}
P.-E. Sarlin, D.~DeTone, T.~Malisiewicz, and A.~Rabinovich, ``Superglue:
  Learning feature matching with graph neural networks,'' in \emph{Proceedings
  of the IEEE/CVF conference on computer vision and pattern recognition}, 2020,
  pp. 4938--4947.

\bibitem{chen2021learning}
H.~Chen, Z.~Luo, J.~Zhang, L.~Zhou, X.~Bai, Z.~Hu, C.-L. Tai, and L.~Quan,
  ``Learning to match features with seeded graph matching network,'' in
  \emph{Proceedings of the IEEE/CVF International Conference on Computer
  Vision}, 2021, pp. 6301--6310.

\bibitem{shi2022clustergnn}
Y.~Shi, J.-X. Cai, Y.~Shavit, T.-J. Mu, W.~Feng, and K.~Zhang, ``Clustergnn:
  Cluster-based coarse-to-fine graph neural network for efficient feature
  matching,'' in \emph{Proceedings of the IEEE/CVF Conference on Computer
  Vision and Pattern Recognition}, 2022, pp. 12\,517--12\,526.

\bibitem{shen2021efficient}
Z.~Shen, M.~Zhang, H.~Zhao, S.~Yi, and H.~Li, ``Efficient attention: Attention
  with linear complexities,'' in \emph{Proceedings of the IEEE/CVF winter
  conference on applications of computer vision}, 2021, pp. 3531--3539.

\bibitem{zhang2018interpretable}
Q.~Zhang, Y.~N. Wu, and S.-C. Zhu, ``Interpretable convolutional neural
  networks,'' in \emph{Proceedings of the IEEE conference on computer vision
  and pattern recognition}, 2018, pp. 8827--8836.

\bibitem{zhang2019interpreting}
Q.~Zhang, Y.~Yang, H.~Ma, and Y.~N. Wu, ``Interpreting cnns via decision
  trees,'' in \emph{Proceedings of the IEEE/CVF conference on computer vision
  and pattern recognition}, 2019, pp. 6261--6270.

\bibitem{williford2020explainable}
J.~R. Williford, B.~B. May, and J.~Byrne, ``Explainable face recognition,'' in
  \emph{European conference on computer vision}.\hskip 1em plus 0.5em minus
  0.4em\relax Springer, 2020, pp. 248--263.

\bibitem{zhao2021towards}
W.~Zhao, Y.~Rao, Z.~Wang, J.~Lu, and J.~Zhou, ``Towards interpretable deep
  metric learning with structural matching,'' in \emph{Proceedings of the
  IEEE/CVF International Conference on Computer Vision}, 2021, pp. 9887--9896.

\bibitem{chen2022guide}
Y.~Chen, D.~Huang, S.~Xu, J.~Liu, and Y.~Liu, ``Guide local feature matching by
  overlap estimation,'' in \emph{Proceedings of the AAAI Conference on
  Artificial Intelligence}, vol.~36, no.~1, 2022, pp. 365--373.

\bibitem{liu2021swin}
Z.~Liu, Y.~Lin, Y.~Cao, H.~Hu, Y.~Wei, Z.~Zhang, S.~Lin, and B.~Guo, ``Swin
  transformer: Hierarchical vision transformer using shifted windows,'' in
  \emph{Proceedings of the IEEE/CVF international conference on computer
  vision}, 2021, pp. 10\,012--10\,022.

\bibitem{yu2023adaptive}
J.~Yu, J.~Chang, J.~He, T.~Zhang, J.~Yu, and F.~Wu, ``Adaptive spot-guided
  transformer for consistent local feature matching,'' in \emph{Proceedings of
  the IEEE/CVF Conference on Computer Vision and Pattern Recognition}, 2023,
  pp. 21\,898--21\,908.

\bibitem{lin2017feature}
T.-Y. Lin, P.~Doll{\'a}r, R.~Girshick, K.~He, B.~Hariharan, and S.~Belongie,
  ``Feature pyramid networks for object detection,'' in \emph{Proceedings of
  the IEEE conference on computer vision and pattern recognition}, 2017, pp.
  2117--2125.

\bibitem{he2016deep}
K.~He, X.~Zhang, S.~Ren, and J.~Sun, ``Deep residual learning for image
  recognition,'' in \emph{Proceedings of the IEEE conference on computer vision
  and pattern recognition}, 2016, pp. 770--778.

\bibitem{8578843}
J.~Hu, L.~Shen, and G.~Sun, ``Squeeze-and-excitation networks,'' in \emph{2018
  IEEE/CVF Conference on Computer Vision and Pattern Recognition}, 2018, pp.
  7132--7141.

\bibitem{cuturi2013sinkhorn}
M.~Cuturi, ``Sinkhorn distances: Lightspeed computation of optimal transport,''
  \emph{Advances in neural information processing systems}, vol.~26, 2013.

\bibitem{balntas2017hpatches}
V.~Balntas, K.~Lenc, A.~Vedaldi, and K.~Mikolajczyk, ``Hpatches: A benchmark
  and evaluation of handcrafted and learned local descriptors,'' in
  \emph{Proceedings of the IEEE conference on computer vision and pattern
  recognition}, 2017, pp. 5173--5182.

\bibitem{truong2023pdc}
P.~Truong, M.~Danelljan, R.~Timofte, and L.~Van~Gool, ``Pdc-net+: Enhanced
  probabilistic dense correspondence network,'' \emph{IEEE Transactions on
  Pattern Analysis and Machine Intelligence}, 2023.

\bibitem{dusmanu2020multi}
M.~Dusmanu, J.~L. Sch{\"o}nberger, and M.~Pollefeys, ``Multi-view optimization
  of local feature geometry,'' in \emph{Computer Vision--ECCV 2020: 16th
  European Conference, Glasgow, UK, August 23--28, 2020, Proceedings, Part I
  16}.\hskip 1em plus 0.5em minus 0.4em\relax Springer, 2020, pp. 670--686.

\bibitem{shen2022semi}
Z.~Shen, J.~Sun, Y.~Wang, X.~He, H.~Bao, and X.~Zhou, ``Semi-dense feature
  matching with transformers and its applications in multiple-view geometry,''
  \emph{IEEE Transactions on Pattern Analysis and Machine Intelligence}, 2022.

\bibitem{rublee2011orb}
E.~Rublee, V.~Rabaud, K.~Konolige, and G.~Bradski, ``Orb: An efficient
  alternative to sift or surf,'' in \emph{2011 International conference on
  computer vision}.\hskip 1em plus 0.5em minus 0.4em\relax Ieee, 2011, pp.
  2564--2571.

\bibitem{bian2017gms}
J.~Bian, W.-Y. Lin, Y.~Matsushita, S.-K. Yeung, T.-D. Nguyen, and M.-M. Cheng,
  ``Gms: Grid-based motion statistics for fast, ultra-robust feature
  correspondence,'' in \emph{Proceedings of the IEEE conference on computer
  vision and pattern recognition}, 2017, pp. 4181--4190.

\bibitem{zhang2019learning}
J.~Zhang, D.~Sun, Z.~Luo, A.~Yao, L.~Zhou, T.~Shen, Y.~Chen, L.~Quan, and
  H.~Liao, ``Learning two-view correspondences and geometry using order-aware
  network,'' in \emph{Proceedings of the IEEE/CVF international conference on
  computer vision}, 2019, pp. 5845--5854.

\bibitem{zhang2021reference}
Z.~Zhang, T.~Sattler, and D.~Scaramuzza, ``Reference pose generation for
  long-term visual localization via learned features and view synthesis,''
  \emph{International Journal of Computer Vision}, vol. 129, no.~4, pp.
  821--844, 2021.

\bibitem{taira2018inloc}
H.~Taira, M.~Okutomi, T.~Sattler, M.~Cimpoi, M.~Pollefeys, J.~Sivic, T.~Pajdla,
  and A.~Torii, ``Inloc: Indoor visual localization with dense matching and
  view synthesis,'' in \emph{Proceedings of the IEEE Conference on Computer
  Vision and Pattern Recognition}, 2018, pp. 7199--7209.

\bibitem{dai2017scannet}
A.~Dai, A.~X. Chang, M.~Savva, M.~Halber, T.~Funkhouser, and M.~Nie{\ss}ner,
  ``Scannet: Richly-annotated 3d reconstructions of indoor scenes,'' in
  \emph{Proceedings of the IEEE conference on computer vision and pattern
  recognition}, 2017, pp. 5828--5839.

\bibitem{lin2014microsoft}
T.-Y. Lin, M.~Maire, S.~Belongie, J.~Hays, P.~Perona, D.~Ramanan,
  P.~Doll{\'a}r, and C.~L. Zitnick, ``Microsoft coco: Common objects in
  context,'' in \emph{European conference on computer vision}.\hskip 1em plus
  0.5em minus 0.4em\relax Springer, 2014, pp. 740--755.

\bibitem{melekhov2020image}
I.~Melekhov, G.~J. Brostow, J.~Kannala, and D.~Turmukhambetov, ``Image
  stylization for robust features,'' \emph{arXiv preprint arXiv:2008.06959},
  2020.

\bibitem{humenberger2020robust}
M.~Humenberger, Y.~Cabon, N.~Guerin, J.~Morat, J.~Revaud, P.~Rerole, N.~Pion,
  C.~de~Souza, V.~Leroy, and G.~Csurka, ``Robust image retrieval-based visual
  localization using kapture,'' \emph{arXiv preprint arXiv:2007.13867}, 2020.

\end{thebibliography}


 




\vfill

\end{document}